\documentclass[11pt]{article}

\usepackage[final]{acl}

\usepackage{times}
\usepackage{latexsym}

\usepackage[T1]{fontenc}

\usepackage[utf8]{inputenc}

\usepackage{microtype}

\usepackage{inconsolata}

\usepackage{graphicx}

\usepackage{tabularx,booktabs}
\usepackage{enumitem}
\usepackage{amsmath}
\usepackage{amssymb}
\usepackage{multirow}
\usepackage{algorithm,algorithmic}
\usepackage{pifont}
\newcommand{\xmark}{\ding{55}}%
\newcommand{\cmark}{\ding{51}}%
\usepackage{colortbl}
\newcommand{\ouralg}{FOREVER}

\usepackage{subcaption}

\title{FOREVER: Forgetting Curve-Inspired Memory Replay for Language Model Continual Learning}

\author{Yujie Feng$^{1}$\thanks{~ Equal contribution.}, Hao Wang$^{2}$\footnotemark[1], Jian Li$^{3}$, Xu Chu$^{4}$, Zhaolu Kang$^{4}$ \\ \textbf{Yiran Liu}$^{5}$, \textbf{Yasha Wang}$^{4}$, \textbf{Philip S. Yu}$^{6}$, \textbf{Xiao-Ming Wu}$^{1}$\\
$^1$The Hong Kong Polytechnic University\enspace
$^2$ShineMo Ltd., China\\
$^3$Solar System of OVB, Tencent, China\enspace
$^4$Peking University \\
$^5$University College London\enspace
$^6$University of Illinois Chicago\\
yujie.feng@connect.polyu.hk, xiao-ming.wu@polyu.edu.hk 
}

\begin{document}
\maketitle
\begin{abstract}

Continual learning (CL) for large language models (LLMs) aims to enable sequential knowledge acquisition without catastrophic forgetting. Memory replay methods are widely used for their practicality and effectiveness, but most rely on fixed, step-based heuristics that often misalign with the model’s actual learning progress, since identical training steps can result in varying degrees of parameter change. Motivated by recent findings that LLM forgetting mirrors the Ebbinghaus human forgetting curve, we propose \textbf{FOREVER} (\textbf{FOR}g\textbf{E}tting cur\textbf{V}e-inspired m\textbf{E}mory \textbf{R}eplay), a novel CL framework that aligns replay schedules with a model-centric notion of time. FOREVER defines model time using the magnitude of optimizer updates, allowing forgetting curve-inspired replay intervals to align with the model’s internal evolution rather than raw training steps. Building on this approach, FOREVER incorporates a forgetting curve-based replay scheduler to determine \emph{when} to replay and an intensity-aware regularization mechanism to adaptively control \emph{how} to replay. Extensive experiments on three CL benchmarks and models ranging from 0.6B to 13B parameters demonstrate that FOREVER consistently mitigates catastrophic forgetting\footnote{The source code is available at \url{https://github.com/WoodScene/FOREVER}}.

\end{abstract}

\section{Introduction}

Enabling large language models (LLMs) with continual learning (CL) capabilities is increasingly important in dynamic environments, where models must acquire new knowledge sequentially without costly retraining over all historical data~\cite{yu2024recent, chang2024survey, eskandar2025star}.
Despite its importance, effective CL for LLMs remains challenging: sequential updates induce distribution drift and interfere with previously learned representations, disrupting the stability–plasticity trade-off and resulting in catastrophic forgetting (CF)~\cite{mccloskey1989catastrophic}.

Replay-based CL methods~\cite{lu2025rethinkingstabilityplasticitytradeoffcontinual, pmlr-v267-wan25d, chen2025prototype, bai2025efficientrehearsalschemecatastrophic}, which revisit a small buffer of past examples while training on new tasks, have emerged as a widely adopted and empirically effective strategy for mitigating forgetting.
However, their effectiveness critically depends on two fundamental design questions:
(i) \textbf{when to replay}, i.e., at which stages of training replay should be activated, and
(ii) \textbf{how to replay}, i.e., how strongly past knowledge should be reinforced relative to current learning.
Existing approaches typically rely on hand-crafted replay heuristics, such as uniform replay intervals or static replay weights, which remain largely decoupled from the model's actual learning dynamics.

\begin{figure}[t]
  \centering
  \includegraphics[width=1\linewidth]{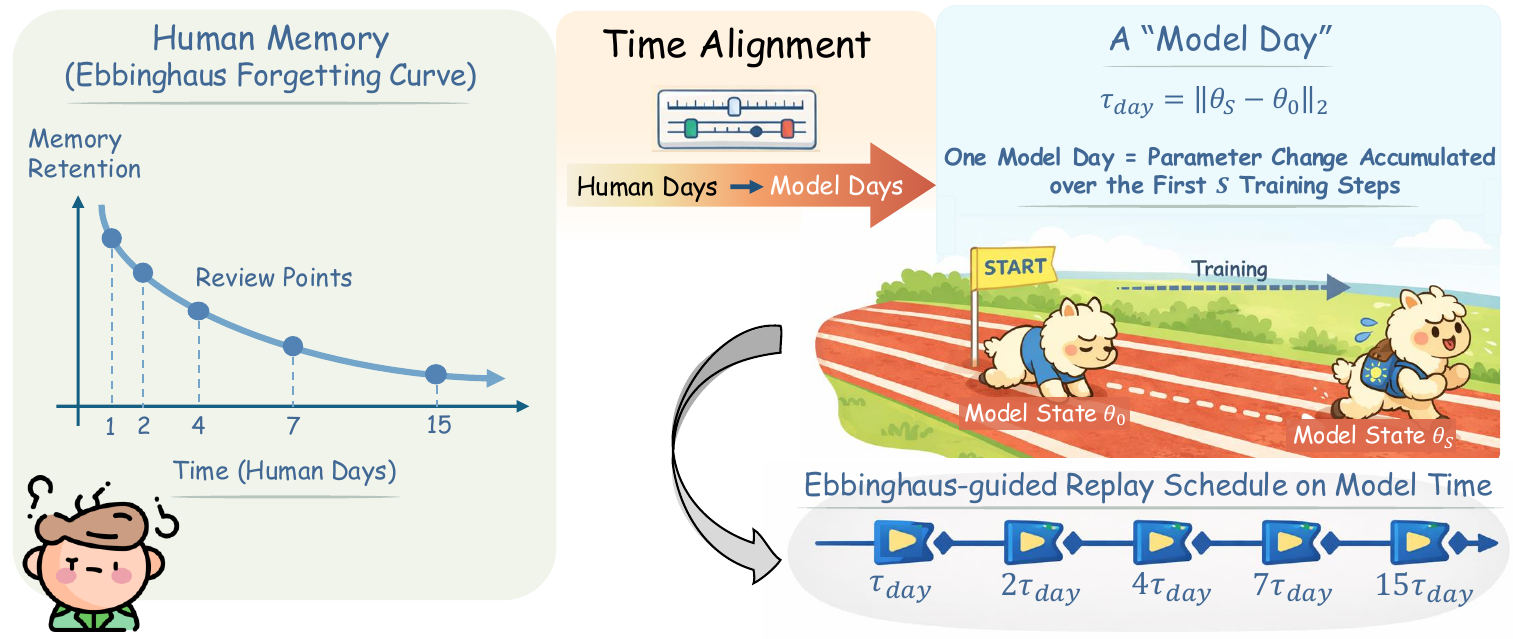}
  \caption{\textbf{Aligning human time and model time in {\ouralg}.}
 {\ouralg} aligns Ebbinghaus-inspired human replay intervals with a model-centric timeline defined by accumulated parameter update magnitude, enabling replay to be triggered based on the model's actual learning progress.
}

  \label{fig:intro}
\end{figure}

Recent studies indicate that neural models exhibit forgetting behaviors similar to those observed in humans~\cite{wu2025humanmemoryaimemory, kline2025human, zhang2025survey, wu2025human}, motivating the application of classical cognitive theories—such as the Ebbinghaus forgetting curve~\cite{murre2015replication}—to LLMs. The Ebbinghaus forgetting curve describes rapid memory loss soon after learning, followed by a slower rate of decay. Drawing on this analogy, recent work has introduced Ebbinghaus-style spaced replay schedules, where replay occurs more frequently shortly after learning and becomes increasingly spaced over time (e.g., 1, 2, 4, 7, 15 days in humans), to better match the temporal dynamics of memory decay~\cite{zhong2024memorybank, chen2025incorporating, kang2025your}.

However, these approaches typically implement replay schedules based on training steps, implicitly equating step count with the passage of human time. This assumption is problematic, as the same number of training steps can result in varying degrees of model change depending on optimization settings such as learning rate or batch size. As a result, step-based replay schedules may trigger replay at inconsistent model states, leading to a misalignment between human time and model time.

This raises a central question:

\textbf{\textit{How can “human days” be aligned with “model days” to enable more reliable replay scheduling?}}

To address this question, we propose a novel continual learning framework, \textbf{{\ouralg}} (\textbf{FOR}g\textbf{E}tting cur\textbf{V}e-inspired m\textbf{E}mory \textbf{R}eplay). {\ouralg} redefines replay scheduling by replacing step-based time with a model-centric notion of time, measured by \textbf{parameter update magnitude}. Unlike training steps, which are external and do not reflect the model's internal changes, parameter update magnitude directly quantifies the extent of model evolution during learning. By tracking this signal, {\ouralg} aligns human-inspired replay schedules with the model's actual update dynamics, calibrating Ebbinghaus-style intervals to the model's learning progress. This ensures that replay is triggered based on true model advancement rather than arbitrary iteration counts.

{\ouralg} consists of two integrated components: an \textbf{forgetting curve-inspired replay scheduler} and an \textbf{intensity-aware replay regularizer}. The scheduler determines \textbf{when to replay} by using accumulated parameter update magnitude to define a model-specific ``day,'' triggering replay at Ebbinghaus-style intervals along this axis. This approach synchronizes replay events with comparable stages of model evolution, effectively mapping the forgetting curve onto the model's learning timeline.

To determine \textbf{how to replay}, {\ouralg} monitors recent update intensity and applies a replay-time regularization term with adaptively adjusted strength. Stronger regularization is used during periods of rapid change to stabilize learning, while gentler regularization is applied as updates slow, supporting new task adaptation. Both replay timing and regularization strength are derived from model update dynamics: accumulated updates indicate progress, and recent intensity reflects the rate of change. This unified approach provides a coherent strategy for both \emph{when} and \emph{how} to replay.

Extensive experiments on diverse CL benchmarks demonstrate that {\ouralg} consistently improves knowledge retention while maintaining strong adaptation to new tasks.

\textbf{Our main contributions are:} \begin{itemize}[leftmargin=*,itemsep=2pt,topsep=0pt,parsep=0pt] \item We introduce {\ouralg}, a novel CL framework that bridges cognitive forgetting theory and model training dynamics by defining time via parameter update magnitude, enabling forgetting curve-inspired replay scheduling beyond step-based heuristics. \item We develop two complementary techniques: an forgetting curve-inspired memory replay scheduler for determining \emph{when} to replay, and an intensity-aware replay regularizer for adaptively controlling \emph{how} to replay. \item Extensive evaluation on three CL benchmarks and four model backbones (0.6B to 13B parameters) demonstrates the effectiveness of {\ouralg} in mitigating catastrophic forgetting. \end{itemize}

\section{Proposed Method: {\ouralg}}
\label{Method}
\begin{figure*}[t]
  \centering
  \includegraphics[width=0.95\linewidth]{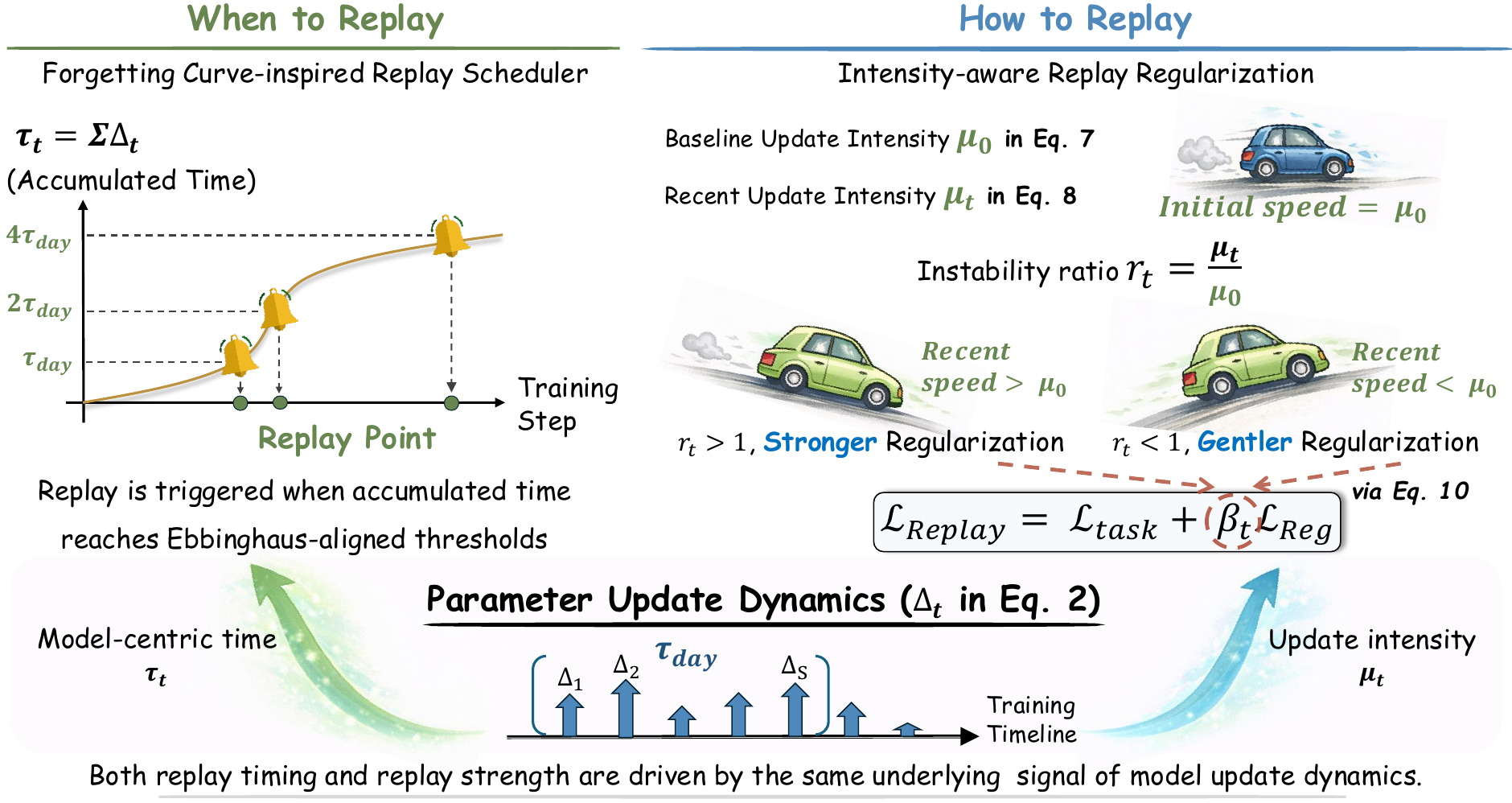}
\caption{\textbf{Overview of {\ouralg}.}
{\ouralg} decomposes replay into two coupled decisions—\emph{when to replay} and \emph{how to replay}—both grounded in model update dynamics.
Parameter update magnitudes $\Delta_t$ track model evolution over training steps, whose accumulation defines a model-centric notion of time (virtual ``model days'').
\textbf{When to replay} (Left): accumulated model time $\tau_t$ measures how far the model has progressed in parameter space and triggers replay when Ebbinghaus-guided time thresholds are reached.
\textbf{How to replay} (Right): recent update intensity $\mu_t$, relative to a baseline $\mu_0$, modulates replay regularization strength—stronger under rapid model changes and gentler when updates are stable.
By unifying replay timing and replay strength under the same update-dynamics signal, {\ouralg} enables a coherent and model-centric replay strategy.
}
  \label{fig:method}
\end{figure*}

\paragraph{Problem Formulation}
Continual learning aims to progressively accumulate knowledge from a sequence of tasks $\{\mathcal{T}_1, \ldots, \mathcal{T}_K\}$ over time. 
Each task $\mathcal{T}_k$ is associated with a dataset $\mathcal{D}_k = \{(x_i^k, y_i^k)\}_{i=1}^{N_k}$ of size $N_k$, where $x_i^k \in \mathcal{X}_k$ denotes the input and $y_i^k \in \mathcal{Y}_k$ denotes the corresponding target.

A model parameterized by $\Theta$ is trained on these tasks sequentially, without access to the full data from previous tasks once they have been observed.
The objective is to minimize the expected negative log-likelihood over the data encountered so far:
\begin{equation}
\mathcal{L} = \mathbb{E}_{(x, y) \sim \mathcal{D}_{\le k}} \left[ -\log p_\Theta(y \mid x) \right],
\end{equation}


Following common practice in replay-based CL, we assume access to a memory buffer that stores a limited number of samples from previous tasks. For each past task $\mathcal{T}_i \ (i < k)$, at most $|\mathcal{M}|$ examples are retained in a task-specific buffer $\mathcal{M}_i$, yielding a global memory $\mathcal{M}_{<k} = \bigcup_{i<k} \mathcal{M}_i$. When learning task $\mathcal{T}_k$, the model is optimized jointly on the current task data $\mathcal{D}_k$ and the replay memory $\mathcal{M}_{<k}$.


\paragraph{Overview}
As illustrated in Figure~\ref{fig:method}, {\ouralg} decomposes replay into two tightly coupled components:
(i) an \textit{\textbf{Forgetting Curve-inspired Replay Scheduler}}, which determines \emph{when to replay} by first calibrating a model-centric notion of time via parameter update dynamics and then mapping human Ebbinghaus intervals onto this model-time axis to trigger replay at the corresponding thresholds; and
(ii) an \textit{\textbf{Intensity-aware Replay Regularization}} mechanism, which regulates \emph{how to replay} by adaptively adjusting replay strength according to recent update intensity.
Together, these components unify replay timing and replay strength under a single dynamics-aware framework, grounding replay decisions in how much and how fast the model evolves, rather than in fixed step-based heuristics.

\subsection{When to Replay: Forgetting Curve-inspired Replay Scheduler}
\paragraph{Model-centric Time Calibration.}
A key challenge in applying Ebbinghaus-style replay schedules to CL lies in defining a meaningful notion of time for LLMs.
While human memory decay is naturally measured in days, training steps are a poor proxy for model progress, as their effects vary substantially with optimization settings such as batch sizes.
To address this mismatch, we introduce a model-centric notion of time grounded in parameter update dynamics, which enables the calibration of \emph{virtual Ebbinghaus days} for LLMs.

\paragraph{Parameter Update Dynamics.}
Let $\Theta_t \in \mathbb{R}^N$ denote the model parameters after the $t$-th training update, where $N$ is the total number of parameters.
We quantify the magnitude of model change at step $t$ using the parameter update norm
\begin{equation}
\Delta_t = \|\Theta_t - \Theta_{t-1}\|_2 ,
\end{equation}
which measures how much the model evolves during a single optimization step.
In practice, $\Delta_t$ is computed over trainable parameters only (i.e., the LoRA~\cite{hu2021loralowrankadaptationlarge} weights) and can be directly obtained from the optimizer's applied parameter updates, without requiring additional forward or backward passes.
Accumulating these updates yields a model-centric notion of time,
\begin{equation}
\tau_t = \sum_{i=1}^{t} \Delta_i ,
\end{equation}
which represents the total distance the model has traveled in parameter space.
In our implementation, $\tau_t$ is reset at the beginning of each new task and measures the elapsed model-centric time since learning on the current task started.
Unlike raw step counts, $\tau_t$ directly reflects model evolution and is substantially less sensitive to optimization hyperparameters.

\paragraph{Calibrating a Virtual Model Day.}
Based on the accumulated update magnitude, we define a virtual model ``day'' as the amount of parameter change observed over an initial window of $S$ training steps:
\begin{equation}
\tau_{\text{day}} = \sum_{i=1}^{S} \Delta_i .
\end{equation}

This quantity serves as a model-specific unit of time, anchoring human-defined temporal intervals to the model's own learning dynamics.

\paragraph{Virtual Ebbinghaus Days.}
Given a sequence of Ebbinghaus-style replay intervals expressed in human days,
$\mathcal{D}_{\text{human}} = \{d_1, d_2, \ldots\}$ (e.g., $\{1, 2, 4, 7, 15, 30, \ldots\}$),
we map them onto the model-centric time axis as
\begin{equation}
\mathcal{D}_{\text{model}} = \{ d \cdot \tau_{\text{day}} \mid d \in \mathcal{D}_{\text{human}} \}.
\end{equation}

Each element in $\mathcal{D}_{\text{model}}$ specifies a target amount of accumulated model change at which replay is triggered, aligning human-inspired replay intervals with comparable stages of model evolution.


The classical Ebbinghaus forgetting curve describes rapid memory decay at early stages, followed by slower decay, motivating replay schedules with increasing intervals.
Leveraging the calibrated model-centric notion of time and replay thresholds $\mathcal{D}_{\text{model}}$, we describe how {\ouralg} determines \emph{when} replay should be triggered.

\paragraph{Replay Triggering Criterion.}
During training, we continuously track the accumulated model-centric time $\tau_t$.
Replay is triggered whenever $\tau_t$ reaches the next threshold in $\mathcal{D}_{\text{model}}$.
Formally, let $j$ denote the index of the next scheduled replay.
A replay event is initiated at training step $t$ if
\begin{equation}
\tau_t \geq \mathcal{D}_{\text{model}}^{(j)}.
\end{equation}
After replay is performed, the scheduler advances to the subsequent threshold, and training resumes on the current task.


\subsection{How to Replay: Intensity-aware Replay Regularization}


Once replay is triggered, {\ouralg} determines how strongly replay should be applied.
The key idea is to adapt replay strength to the model's current training dynamics: replay should impose stronger constraints when the model is changing rapidly and remain gentle once learning stabilizes.
To this end, {\ouralg} modulates replay regularization based on the ratio between recent and baseline update intensity over training.

\paragraph{Update Intensity and Instability Ratio.}
Using the same warm-up window of $S$ steps, we define the baseline update intensity as
\begin{equation}
\mu_0 = \frac{1}{S} \sum_{t=1}^{S} \Delta_t ,
\end{equation}
which provides a reference scale for subsequent training dynamics.

Thereafter, the recent update intensity is tracked using an exponential moving average,
\begin{equation}
\mu_t = (1-\lambda)\mu_{t-1} + \lambda \Delta_t ,
\label{eq:lambda}
\end{equation}
where $\lambda$ controls the degree of temporal smoothing.
We then compute an instability ratio
\begin{equation}
r_t = \frac{\mu_t}{\mu_0},
\end{equation}
which indicates whether the model is currently updating more aggressively ($r_t > 1$) or more conservatively ($r_t < 1$) than at the beginning of training.

The replay regularization strength is scaled according to the instability ratio $r_t$:
\begin{equation}
\beta_t
=
\beta_{\text{base}}
\cdot
\mathrm{clip}
\!\left(
1 + \gamma (r_t - 1),
\, g_{\min},\, g_{\max}
\right),
\label{eq:gbeta}
\end{equation}
where $\beta_{\text{base}}$ balances the magnitude between the replay regularization loss and the task loss, $\gamma$ controls sensitivity to update intensity, and $g_{\min}$ and $g_{\max}$ denote the lower and upper bounds of the clipping operation for numerical stability.


\paragraph{Replay Objective.}
At replay time, the model is optimized on samples from the memory buffer while applying a parameter-level regularization anchored at a reference snapshot $\Theta^\star$, taken at the end of the previous task.
The replay loss is defined as
\begin{equation}
\mathcal{L}_{\text{replay}}
=
\mathcal{L}_{\text{task}}^{(\text{old})}
+
\beta_t
\sum_{j}
\|\Theta_j - \Theta_j^\star\|_2^2,
\label{eq:loss}
\end{equation}
where $\theta_j$ denotes an individual model parameter.

By scaling replay regularization with the instability ratio $r_t$, {\ouralg} applies stronger constraints when the model undergoes rapid changes and relaxes replay as learning stabilizes.
This design enables adaptive replay strength driven solely by parameter update dynamics, without relying on fixed replay weights or hand-crafted schedules.


Detailed implementation of {\ouralg} algorithm is provided in the Appendix (Algorithm~\ref{alg:my_algorithm}).

\section{Experiments and Analysis}\label{sec:exp}

\paragraph{Datasets}
Following the experimental protocol of \citet{du2024unlocking}, we evaluate on three representative CL benchmarks for NLP:
(i) the \textbf{Standard CL Benchmark}, which comprises five text classification tasks from \citet{zhang2015character};
(ii) the \textbf{Long Sequence Benchmark}~\cite{razdaibiedina2023progressive}, a more challenging setting consisting of 15 sequential tasks designed to assess long-horizon knowledge accumulation and retention; and
(iii) the \textbf{SuperNI Benchmark}~\cite{wang2022super}, a large-scale instruction-following benchmark containing 15 diverse NLP generation tasks.
Following \citet{wang2023orthogonal}, we sample 1000 training instances per task and reserve 500 instances per class for evaluation.
For each benchmark, we evaluate multiple task orders, with detailed dataset statistics and task sequences provided in Appendix~\ref{sec:dataset}.

\newcommand{\tabincell}[2]{\begin{tabular}{@{}#1@{}}#2\end{tabular}}
\begin{table*}[t]
\centering
\scalebox{0.95}{
\begin{tabular}{l|c|cccccc}
\toprule
\multirow{2}*{\tabincell{c}{Method}}  & \multirow{2}*{\tabincell{c}{Memory-Based}}  & \multicolumn{2}{c}{Standard CL} & \multicolumn{2}{c}{Long Sequence } & \multicolumn{2}{c}{SuperNI }\\
  & & OP$\uparrow$ &   BWT$\uparrow$ & OP$\uparrow$&   BWT$\uparrow$  & OP$\uparrow$&   BWT$\uparrow$ \\
\midrule
\rule{0pt}{4pt}Fine-tuning  &  \multirow{4}*{\tabincell{c}{\xmark}} & 47.2 & -12.6 & 36.0 & -17.5 & 8.2 & -27.4  \\
\rule{0pt}{8pt}EWC~\cite{kirkpatrick2017overcoming}& & 51.0 & -10.3 & 44.8 & -13.8 & 32.9 & -18.6  \\
\rule{0pt}{8pt}O-LoRA~\cite{wang2023orthogonal} & &59.4 & -7.9 & 54.1 & -12.4 & 23.7 & -17.5  \\
\rule{0pt}{8pt}MoELoRA~\cite{luo2024moelora} && 55.3 & -8.2 & 35.3 & -13.6 & 25.7 & -11.3 \\

\midrule

\rule{0pt}{8pt}MixReplay  &  \multirow{9}*{\tabincell{c}{\cmark}}  & 65.8 & -8.0 & 65.1 & -11.4 & 34.6 & -14.1 \\
\rule{0pt}{8pt}Fixed-interval Replay  && 65.1 & -9.2 & 64.5  & -10.9  & 34.7 & -14.5\\
\rule{0pt}{8pt}SAPT~\cite{zhao2024sapt} & & 68.8 & -6.9 & 67.2 & -8.8 & 38.5 & -6.2 \\
\rule{0pt}{8pt}MIGU~\cite{du2024unlocking} && 69.9 & -7.5 & 66.3 & -9.2 & 35.0 & -8.1 \\
\rule{0pt}{8pt}SSR~\cite{huang2024mitigating}& & 68.4 & -7.1 & 67.5 & -9.0 & 40.1 & -5.4 \\
\rule{0pt}{8pt}Recurrent-KIF~\cite{feng2025recurrent} && 70.6 & -6.5 & 67.7 & -8.4 & 39.7 & -5.0  \\
\rule{0pt}{8pt}AIMMerging~\cite{feng2025aimmerging} && 71.9 & -5.0 & 67.9 & -6.3 & 41.0 & -3.4 \\
\rule{0pt}{8pt}VBM~\cite{kang2025your} && 71.5 & -5.2 & 68.1 & -6.1 & 41.3 & -3.7 \\

\rowcolor[gray]{0.9}
\rule{0pt}{8pt}\textbf{{\ouralg} (ours)} && \textbf{72.9} & \textbf{-4.7} & \textbf{69.4} & \textbf{-5.0} & \textbf{42.1} & \textbf{-2.9} \\

\midrule

\rule{0pt}{8pt} MTL  & & 77.4  & - & 77.8  & -  & 48.2 & - \\
\bottomrule
\end{tabular}}
\caption{Overall CL results on three benchmarks using the Qwen3-0.6B backbone.
We report Overall Performance (OP) and Backward Transfer (BWT) after training on the final task.
All results are averaged over different task orders.
The last row corresponds to the multi-task learning (MTL) upper bound.
Our method, {\ouralg}, outperforms the previous best method, VBM, with an average improvement of 1.2\% in OP and a 0.9\% increase in BWT.
}

\label{tbl:result}
\end{table*}

\paragraph{Metrics}
Let $a_{i,j}$ denote the testing performance on task $\mathcal{T}_i$ after training on task $\mathcal{T}_j$, and let $K$ denote the total number of tasks.
We evaluate the overall performance (OP) \cite{chaudhry2018riemannian} and backward transfer (BWT) \cite{ke2022continual} after training on the final task:
\begin{equation}
    \mathbf{OP} =\frac{1}{K} \sum_{i=1}^{K} a_{i, K}
\end{equation}
\begin{equation}
    \mathbf{BWT} = \frac{1}{K-1} \sum\limits_{i=1}^{K-1} (a_{i, K}-a_{i, i})
\end{equation}

\paragraph{Baselines}
We compare {\ouralg} with a comprehensive set of CL baselines, with a particular emphasis on replay-based methods that adopt different replay scheduling strategies.
For fairness, all methods are implemented under the same LoRA-based framework and trained for the same number of epochs.
The evaluated baselines include MixReplay, Fixed-interval Replay, EWC~\cite{kirkpatrick2017overcoming}, O-LoRA~\cite{wang2023orthogonal}, MoELoRA~\cite{luo2024moelora}, SAPT~\cite{zhao2024sapt}, MIGU~\cite{du2024unlocking}, SSR~\cite{huang2024mitigating}, Recurrent-KIF~\cite{feng2025recurrent}, AIMMerging~\cite{feng2025aimmerging}, and VBM~\cite{kang2025your}.
Finally, multi-task learning (MTL), which jointly trains on all tasks, serves as an upper-bound reference.
Detailed descriptions of all baselines are provided in Appendix~\ref{appendix:baseline}.

\paragraph{Training Details}
We evaluate {\ouralg} across multiple backbone models, including Qwen3-0.6B and Qwen3-4B~\cite{qwen3}, as well as LLaMA3.1-8B~\cite{touvron2023llama} and LLaMA2-13B.
Following \citet{feng2025recurrent}, we store 2\% of the original training data from each task in a memory buffer for replay.
For {\ouralg}, the warm-up window size $S$ is set to 24, which is used to calibrate both the virtual model day and the baseline update intensity.
The smoothing coefficient $\lambda$ in Eq.~(\ref{eq:lambda}) is fixed to 0.05.
The base regularization coefficient $\beta_{\text{base}}$ in Eq.~(\ref{eq:gbeta}) is set to $10^{-3}$, with clipping bounds $g_{\min}=0.5$ and $g_{\max}=3.0$.
All experiments are averaged over three independent runs.
More details are provided in Appendix \ref{sec:details}.

\subsection{Main Results}

The overall CL results using the same Qwen3-0.6B backbone are summarized in Table~\ref{tbl:result}.

\paragraph{{\ouralg} Effectively Mitigates the Challenge of Catastrophic Forgetting.}

{\ouralg} consistently outperforms representative CL baselines, including regularization-based methods (e.g., EWC) and replay-based approaches (e.g., SSR and Recurrent-KIF).
In particular, {\ouralg} achieves a notable improvement in OP, increasing the average OP from 58.7\% to 61.5\% compared to the strongest replay-based baseline, SSR.
Meanwhile, {\ouralg} exhibits stronger backward transfer, improving BWT from -4.9\% to -4.2\% relative to AIMMerging.
These results indicate that {\ouralg} more effectively preserves knowledge from earlier tasks while maintaining competitive performance on newly learned tasks.

Moreover, {\ouralg} consistently surpasses the Ebbinghaus-inspired VBM baseline, yielding gains of 1.2\% in OP and 0.9\% in BWT.
This demonstrates the advantage of aligning replay decisions with model-centric learning dynamics, enabling more robust knowledge retention than step-based replay or static regularization strategies.

\begin{figure}[t]
  \centering
  \includegraphics[width=1\linewidth]{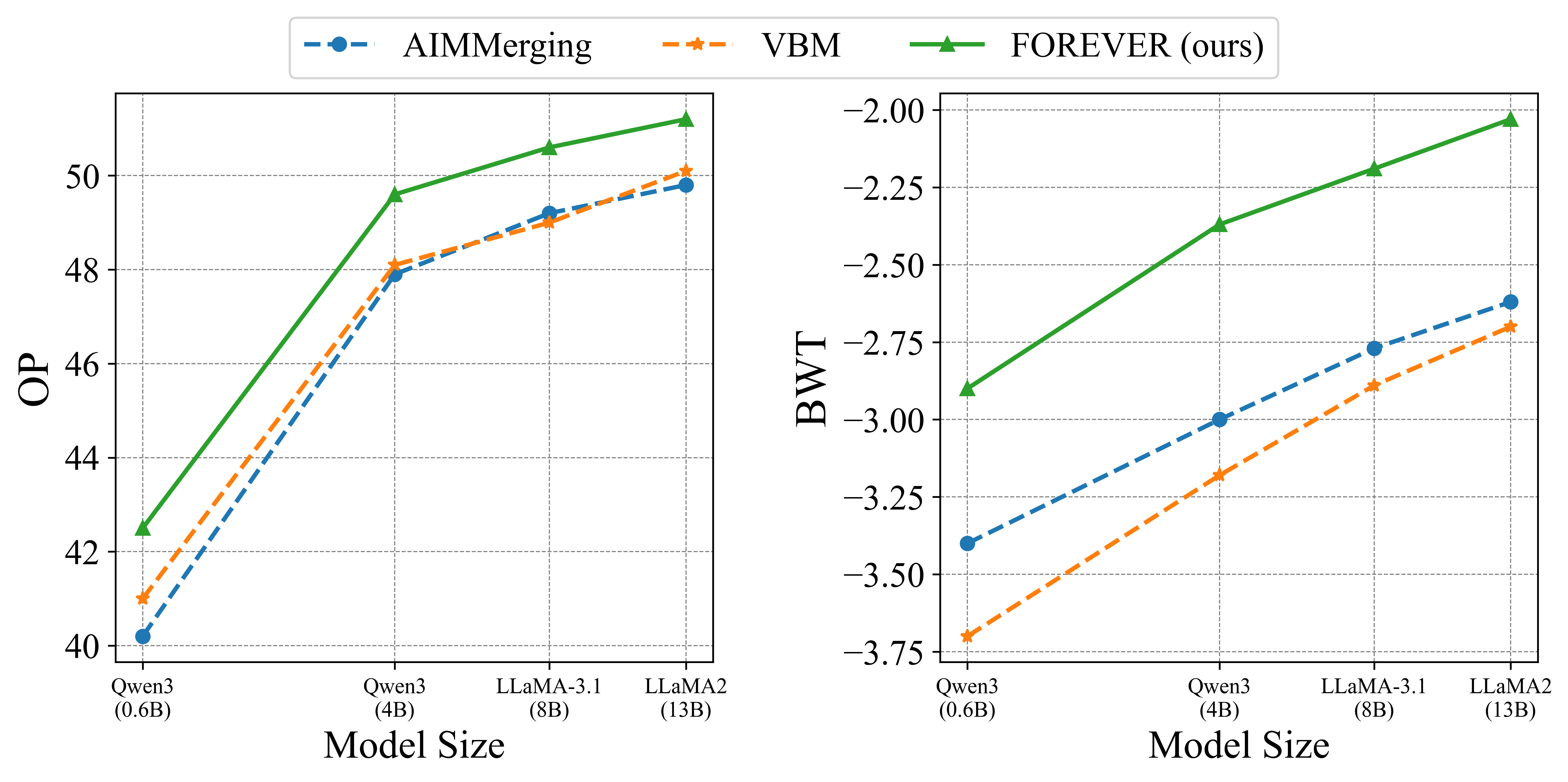}
  \caption{Performance of {\ouralg} with different backbones on the SuperNI Benchmark.}
  \label{fig:different_size}
\end{figure}

\paragraph{{\ouralg} Generalizes Consistently Across Model Scales.}
We further evaluate {\ouralg} across backbone models ranging from 0.6B to 13B parameters, as shown in Figure~\ref{fig:different_size}.
Across all model scales, {\ouralg} consistently outperforms the corresponding baselines.
For example, with the LLaMA3.1-8B backbone, {\ouralg} improves OP from 49.0\% to 50.6\% and reduces BWT from $-2.9\%$ to $-2.1\%$ compared to VBM.
These results demonstrate that the benefits of model-centric replay scheduling generalize across backbone sizes.

\begin{figure}[t]
  \centering
  \includegraphics[width=1\linewidth]{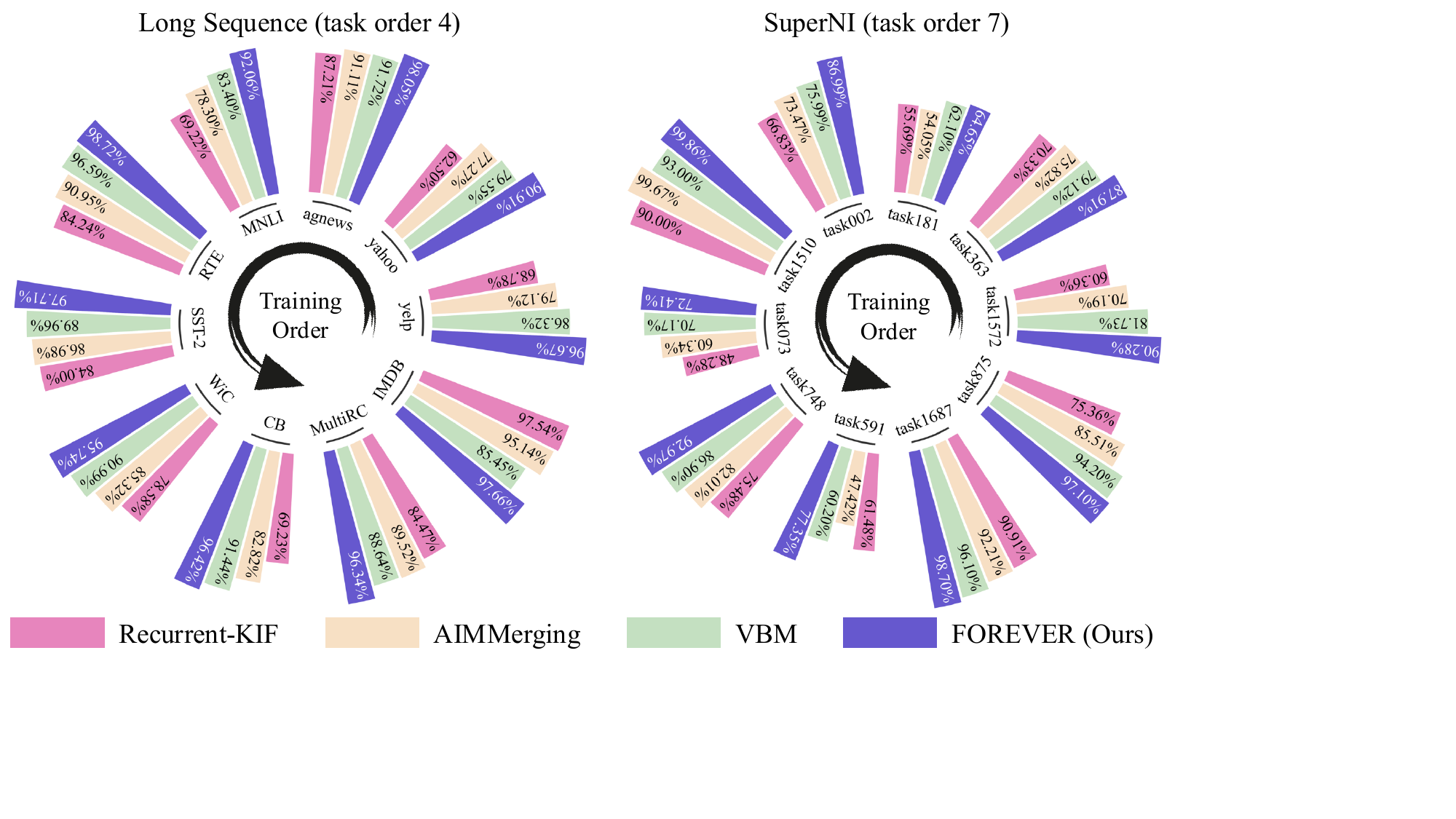}
  \caption{\textbf{Impact of Catastrophic Forgetting in Continual Learning.}
  Performance is reported as a percentage of each task's upper bound.
  After fine-tuning on the final task, {\ouralg} shows superior resistance to performance decline on previously learned tasks.
  }
  \label{fig:radial_plot}
\end{figure}

\paragraph{Forgetting Curve-inspired Replay Enables Knowledge Retention.}
Figure~\ref{fig:radial_plot} presents a task-wise comparison after training on the final task.
{\ouralg} consistently retains higher performance on previously learned tasks than all baselines.
Notably, on RTE and task1687, {\ouralg} achieves performance comparable to the multi-task learning upper bound, indicating strong resistance to catastrophic forgetting.
Overall, these results suggest that {\ouralg} balances prior knowledge retention and new task adaptation in continual learning.

\subsection{Ablation Study}
We conduct ablation studies to analyze the contribution of individual components in {\ouralg}.
Results for task order 7 on the SuperNI are reported in Table~\ref{tbl:ablation}.
Additional analysis, including time complexity, memory size sensitivity, and hyperparameter robustness, are provided in Appendix~\ref{sec:hyper}.

\subsubsection{Effect of the Forgetting Curve-inspired Replay Scheduler}

We first examine how different replay schedules affect CL performance to validate the Ebbinghaus-style design.
Specifically, we replace the human-derived schedule $\mathcal{D}_{\text{human}}$ with alternative strategies to construct model-driven replay sequences $\mathcal{D}_{\text{model}}$:
(i) \textit{Fixed-interval replay} (``+FIR''), which triggers replay at uniform intervals (e.g., $\{24, 48, 96, \ldots\}$);
(ii) \textit{Reversed replay} (``+RR''), which applies replay intervals in descending order (e.g., $\mathcal{D}_{\text{human}} = \{30, 15, 7, 4, 2, 1\}$);
and (iii) \textit{End-only replay} (``+ER''), which performs replay only after completing training on the new task.

Across all variants, the forgetting curve-inspired schedule consistently yields the strongest performance.
This result indicates that replaying more frequently during early training—when parameter updates are large—and gradually reducing replay frequency as learning stabilizes is critical for mitigating catastrophic forgetting.

\subsubsection{Effect of Model-Centric Time Calibration}
We next investigate the impact of aligning human time with model-centric time on replay effectiveness.
We compare our update-dynamics-based calibration with a \textit{step-based calibration} baseline (``+STC''), which maps human-defined days to fixed training steps.

Model-centric calibration consistently outperforms step-based calibration, yielding an average improvement of 1.2\% in OP and 1.1\% in BWT.
This gap arises because step-based schedules trigger replay at fixed iteration counts across tasks, ignoring task-specific learning dynamics.
In contrast, our approach adapts replay timing to the model's intrinsic evolution, enabling more accurate alignment between replay events and learning stages.



\begin{table}[t]
\centering
\scalebox{0.8}{
\begin{tabular}{clcc}
\toprule
Category & Method & OP $\uparrow$ & BWT $\uparrow$ \\
\midrule

\rowcolor[gray]{0.95}
\multirow{1}{*}{Full Model}
& {\ouralg} & 42.5 & -2.8
 \\

\midrule
\multirow{3}{*}{\shortstack{Replay Scheduler\\ (\S 3.2.1)}}

& \hspace{0.4cm} $+$ FIR & 40.1 & -5.2    \\
& \hspace{0.4cm} $+$ RR   & 37.2 & -7.8 \\
& \hspace{0.4cm} $+$ ER  & 40.9 & -6.9 \\

\midrule

Time Calibration \\ (\S 3.2.2)& \hspace{0.4cm} $+$ STC    & 41.3 & -3.9    \\

\midrule
\multirow{3}{*}{\shortstack{Regularization\\ (\S 3.2.3)}}
& \hspace{0.4cm} $-$ IAR    & 39.9 & -4.4    \\
& \hspace{0.4cm} $+$ PIR     & 42.7 & -3.0   \\
& \hspace{0.4cm} $+$ IAR \& PIR  & 42.8 & -2.6       \\

\bottomrule
\end{tabular}}
\caption{
\textbf{Ablation study.}
We evaluate the contribution of each component by selectively removing (``$-$'') or replacing (``$+$'') it with alternative designs.
}

\label{tbl:ablation}
\end{table}

\subsubsection{Effect of Intensity-aware Replay Regularization}
Finally, we evaluate the role of intensity-aware replay regularization in {\ouralg}.
We compare our approach with three variants:
(i) removing the regularization term (``$-$IAR'');
(ii) replacing it with parameter-importance regularization following EWC (``+PIR''); and
(iii) combining importance with intensity-aware scaling (``+IAR \& PIR'').

Removing IAR leads to clear performance drops, confirming its necessity.
Compared with PIR, {\ouralg} achieves comparable results while avoiding the additional cost of estimating and storing parameter importance scores.
Combining IAR and PIR yields only marginal gains, suggesting that both mechanisms address similar sources of instability during training.
Overall, these results demonstrate that parameter update dynamics provide an efficient and effective signal for modulating replay strength.


\subsection{Visualization of Replay Dynamics}

\begin{figure}[t]
  \centering
  \includegraphics[width=1\linewidth]{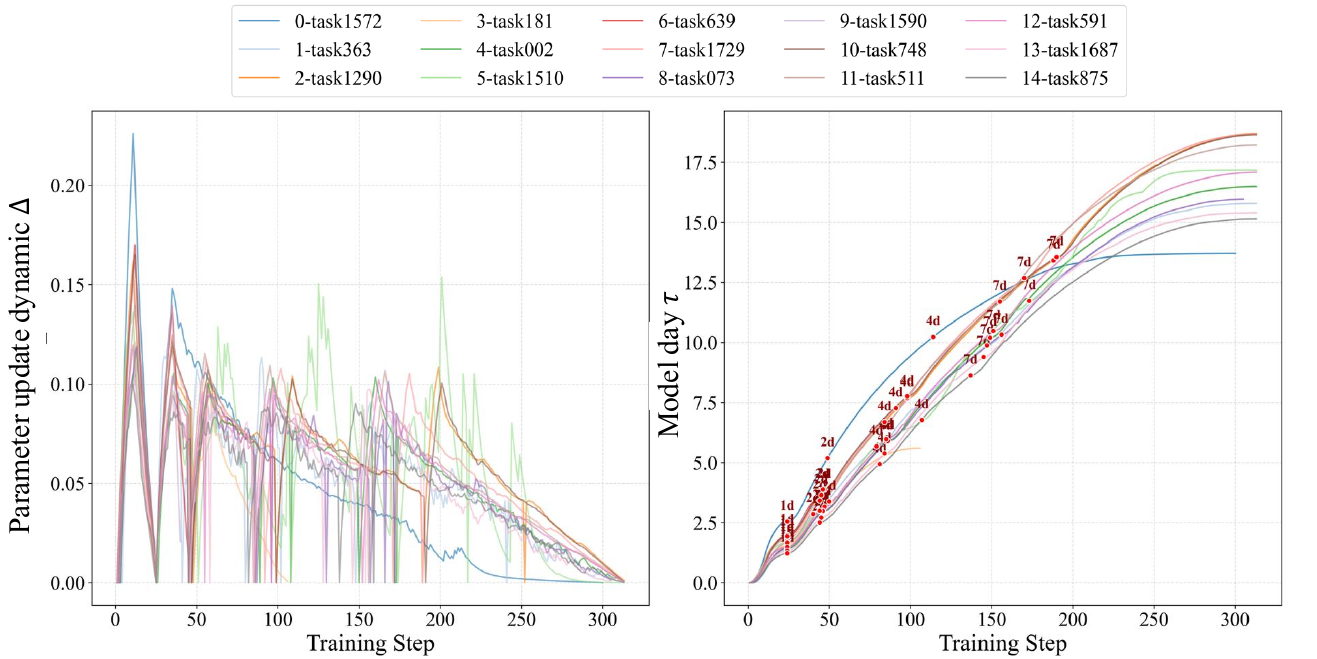}
\caption{
\textbf{Visualization of model-centric replay dynamics during training.}
\textbf{Left:} step-wise parameter update magnitude $\Delta_t$ across training steps.
\textbf{Right:} accumulated model-centric time $\tau_t$ with replay trigger points annotated.
Under the proposed model-centric time definition, replay is triggered at different training steps for different tasks, reflecting task-dependent learning dynamics rather than fixed step-based schedules.
}

  \label{fig:Vis_main}
\end{figure}

We visualize the replay dynamics induced by the proposed model-centric scheduling mechanism.
Specifically, we analyze the step-wise parameter update magnitude $\Delta_t$, the evolution of accumulated model-centric time $\tau_t$, and the resulting replay trigger points throughout training.
Figure~\ref{fig:Vis_main} shows results for task order 7 on the SuperNI benchmark, while visualizations for all eight task orders are provided in Appendix~\ref{appendix:vis} (Figure~\ref{fig:8sub}).

As shown in the left panel, the update magnitude $\Delta_t$ exhibits pronounced non-uniformity over training.
Large updates occur at the early stages of each task, followed by progressively smaller updates as optimization stabilizes.
Correspondingly, the accumulated model-centric time $\tau_t$ (right panel) grows rapidly during periods of substantial parameter change and increases more slowly once training enters a stable regime.

Importantly, under {\ouralg}, replay events are not tied to fixed training steps.
Instead, replay is triggered whenever $\tau_t$ reaches the next Ebbinghaus-calibrated threshold.
As a result, the training steps at which replay occurs vary substantially across tasks.
For example, the replay corresponding to the ``7-day'' threshold spans a wide range of training steps (approximately steps 140--180) across different tasks.
This behavior illustrates that replay in {\ouralg} is governed by the model's intrinsic learning progress rather than by predefined step-based schedules, providing a direct explanation for its improved robustness to catastrophic forgetting.

\subsection{Effect of Increasing-Spacing Replay Scheduling}
\label{sec:replay_schedule}

We further analyze the effect of replay interval scheduling in continual learning. 
The key idea behind our design is not a specific handcrafted interval sequence, 
but a structural principle: forgetting typically follows a fast-then-slow decay pattern, 
which motivates a dense-to-sparse (increasing-spacing) replay schedule. 
In early training stages, frequent reinforcement is required, while later stages allow progressively wider intervals.

To isolate the effect of interval structure, we conduct controlled comparisons under identical replay budgets and cumulative replay data volumes. 
Our original setting adopts an Ebbinghaus-inspired increasing-spacing schedule as the default configuration. 
We additionally consider several alternative scheduling strategies:

\begin{itemize}
[leftmargin=*,itemsep=2pt,topsep=0pt,parsep=0pt] 
    \item \textbf{Increasing-spacing}: including multiple parameterizations, such as exponential (geometric progression) and polynomial (square growth);
    \item \textbf{Uniform-spacing}: equal interval replay;
    \item \textbf{Decreasing-spacing}: reversed scheduling (sparse-to-dense).
\end{itemize}

Results on SuperNI (Qwen3-0.6B, task order 7) are shown in Table~\ref{tab:replay_schedule}. 
We observe that increasing-spacing schedules consistently outperform uniform and decreasing-spacing strategies. 
This aligns with the intuition that replay frequency should match the evolving forgetting dynamics: frequent replay is needed when forgetting is rapid, while wider intervals suffice when forgetting slows down.

Among different increasing-spacing variants, the standard Ebbinghaus-style instantiation achieves the best performance. 
Notably, this schedule is not an arbitrary functional choice, but is grounded in long-established empirical studies of human memory. 
Compared with purely parametric forms (e.g., exponential or polynomial growth), it reflects a more refined spacing pattern.

Overall, these results suggest that performance gains stem from alignment with the underlying forgetting behavior, rather than any specific handcrafted interval sequence.

\begin{table*}[t]
\centering
\small
\begin{tabular}{lccc}
\toprule
\textbf{Replay Strategy} & $D_{\text{human}}$ & OP $\uparrow$ & BWT $\uparrow$ \\
\midrule
Increasing-spacing (standard) & $\{1,2,4,7,15,\dots\}$ & 42.5 & -2.8 \\
Increasing-spacing (exponential) & $\{1,2,4,8,16,\dots\}$ & 42.3 & -2.6 \\
Increasing-spacing (polynomial) & $\{1,4,9,16,\dots\}$ & 41.5 & -3.2 \\
Uniform-spacing & $\{2,4,6,8,10,\dots\}$ & 40.9 & -3.6 \\
Decreasing-spacing & $\{15,7,4,2,1\}$ & 37.2 & -7.8 \\
\bottomrule
\end{tabular}
\caption{Comparison of different replay interval schedules under the same replay budget.}
\label{tab:replay_schedule}
\end{table*}

\section{Related Work}

\subsection{Continual Learning for LLMs}

Continual learning (CL) aims to enable models to acquire new knowledge while mitigating catastrophic forgetting of previously learned information~\cite{ke2021achieving, zhou2024continual, jiang2025unlocking}.
Extending CL to LLMs has recently gained traction, as LLMs are expected to adapt to evolving tasks without full retraining~\cite{pham2023continual, zeng-etal-2025-task, wang2025see}.

Existing CL approaches for LLMs fall into three major categories:
(1) Regularization-based methods constrain parameter updates based on estimated importance scores~\cite{li2024revisiting, zhang2025merge, cheng2025continuous, momeni2025anacp}, but their scalability is challenged by the prohibitive cost of importance estimation in LLMs~\cite{tang2025mergingmodelsflyretraining, ong2025towards}.
(2) Architecture-based methods allocate task-specific components such as adapters, orthogonal subspaces, or MoE modules~\cite{ke2023sub, wan2024knowledge, he2024seekr, wang2024self, feng2024tasl}, which reduce interference but often introduce additional memory or require task identifiers, limiting scalability.
(3) Replay-based methods revisit selected samples from previous tasks~\cite{huang2024mitigating, alexandrov2024mitigating}, and have shown strong empirical performance when combined with PEFT techniques like LoRA~\cite{huang2025mitigating}.

However, most replay-based methods rely on heuristic schedules—such as fixed intervals or mixing ratios—that are decoupled from the model's learning dynamics~\cite{wan2025empowering}.
This raises a central question: \emph{when and how should replay be triggered to align with the model's internal state?}
Our work addresses this gap by grounding replay decisions in the model's intrinsic learning dynamics, moving beyond step-based or static schedules.


\subsection{Forgetting Dynamics in LLMs}
The forgetting behavior of LLMs has recently gained attention from perspectives such as representation drift, knowledge loss, and long-term memory degradation~\cite{jiang2024interpretable, dou-etal-2024-loramoe, ren2024analyzing, li2025dynamic, liu2025advances}. Beyond CL, forgetting is now studied as a general property of LLMs under sequential or prolonged training~\cite{zeng2025modalprompt, wang2025recall}.

Recent empirical studies show that forgetting in LLMs exhibits structured temporal patterns resembling human memory decay, with rapid early performance drops followed by slower degradation~\cite{wu2025humanmemoryaimemory, kline2025human}.
This has motivated the adoption of cognitive principles such as the Ebbinghaus forgetting curve in CL, where replay intervals increase over training~\cite{deng2025unlockingpowerrehearsalcontinual}.
However, existing approaches typically measure ``time'' using training steps, implicitly assuming that iteration counts reflect comparable model changes—an assumption that often breaks down under the highly variable optimization dynamics of LLMs.
In contrast, our work grounds Ebbinghaus-style replay in a model-centric notion of time defined by parameter update dynamics, enabling replay decisions aligned with the model's intrinsic learning process.



\section{Conclusion}

In this paper, we propose {\ouralg}, a replay-based continual learning framework for large language models that aligns human-inspired replay schedules with a model-centric notion of time.
By grounding replay decisions in parameter update dynamics, {\ouralg} jointly determines when to replay and how strongly to regularize past knowledge.
Extensive experiments across multiple benchmarks and model scales show that {\ouralg} effectively mitigates catastrophic forgetting and consistently outperforms existing continual learning methods.


\section*{Limitations}

We acknowledge two primary limitations of our work, each of which highlights promising avenues for future research.

First, {\ouralg} utilizes parameter update dynamics as an indirect proxy for model evolution and forgetting. While metrics such as accumulated update magnitude and recent update intensity offer a principled, model-centric notion of time, they do not directly reflect task-level performance degradation or semantic forgetting. Incorporating more explicit indicators of forgetting—such as performance-based metrics or task-specific diagnostics—could further enhance the effectiveness of replay scheduling.

Second, {\ouralg} employs a predefined Ebbinghaus-inspired replay interval pattern, drawing from cognitive theory. Although this approach provides an interpretable and effective prior for replay, it may not be universally optimal across diverse tasks or training settings. Learning replay interval patterns from data, or dynamically adapting them to specific task characteristics, represents a valuable direction for extending beyond fixed, human-inspired schedules.


\bibliography{acl_latex}

@article{wan2024knowledge,
  title={Knowledge fusion of large language models},
  author={Wan, Fanqi and Huang, Xinting and Cai, Deng and Quan, Xiaojun and Bi, Wei and Shi, Shuming},
  journal={arXiv preprint arXiv:2401.10491},
  year={2024}
}

@inproceedings{chaudhry2018riemannian,
  title={Riemannian walk for incremental learning: Understanding forgetting and intransigence},
  author={Chaudhry, Arslan and Dokania, Puneet K and Ajanthan, Thalaiyasingam and Torr, Philip HS},
  booktitle={Proceedings of the European conference on computer vision (ECCV)},
  pages={532--547},
  year={2018}
}

@article{ke2022continual,
  title={Continual learning of natural language processing tasks: A survey},
  author={Ke, Zixuan and Liu, Bing},
  journal={arXiv preprint arXiv:2211.12701},
  year={2022}
}

@article{kirkpatrick2017overcoming,
  title={Overcoming catastrophic forgetting in neural networks},
  author={Kirkpatrick, James and Pascanu, Razvan and Rabinowitz, Neil and Veness, Joel and Desjardins, Guillaume and Rusu, Andrei A and Milan, Kieran and Quan, John and Ramalho, Tiago and Grabska-Barwinska, Agnieszka and others},
  journal={Proceedings of the national academy of sciences},
  volume={114},
  number={13},
  pages={3521--3526},
  year={2017},
  publisher={National Acad Sciences}
}

@article{luo2024moelora,
  title={Moelora: Contrastive learning guided mixture of experts on parameter-efficient fine-tuning for large language models},
  author={Luo, Tongxu and Lei, Jiahe and Lei, Fangyu and Liu, Weihao and He, Shizhu and Zhao, Jun and Liu, Kang},
  journal={arXiv preprint arXiv:2402.12851},
  year={2024}
}

@inproceedings{zhao2024sapt,
  title={Sapt: A shared attention framework for parameter-efficient continual learning of large language models},
  author={Zhao, Weixiang and Wang, Shilong and Hu, Yulin and Zhao, Yanyan and Qin, Bing and Zhang, Xuanyu and Yang, Qing and Xu, Dongliang and Che, Wanxiang},
  booktitle={Proceedings of the 62nd Annual Meeting of the Association for Computational Linguistics (Volume 1: Long Papers)},
  pages={11641--11661},
  year={2024}
}

@article{touvron2023llama,
  title={Llama: Open and efficient foundation language models},
  author={Touvron, Hugo and Lavril, Thibaut and Izacard, Gautier and Martinet, Xavier and Lachaux, Marie-Anne and Lacroix, Timoth{\'e}e and Rozi{\`e}re, Baptiste and Goyal, Naman and Hambro, Eric and Azhar, Faisal and others},
  journal={arXiv preprint arXiv:2302.13971},
  year={2023}
}

@article{qwen3,
    title={Qwen3 Technical Report}, 
    author={An Yang and Anfeng Li and Baosong Yang and Beichen Zhang and Binyuan Hui and Bo Zheng and Bowen Yu and Chang Gao and Chengen Huang and Chenxu Lv and Chujie Zheng and Dayiheng Liu and Fan Zhou and Fei Huang and Feng Hu and Hao Ge and Haoran Wei and Huan Lin and Jialong Tang and Jian Yang and Jianhong Tu and Jianwei Zhang and Jianxin Yang and Jiaxi Yang and Jing Zhou and Jingren Zhou and Junyang Lin and Kai Dang and Keqin Bao and Kexin Yang and Le Yu and Lianghao Deng and Mei Li and Mingfeng Xue and Mingze Li and Pei Zhang and Peng Wang and Qin Zhu and Rui Men and Ruize Gao and Shixuan Liu and Shuang Luo and Tianhao Li and Tianyi Tang and Wenbiao Yin and Xingzhang Ren and Xinyu Wang and Xinyu Zhang and Xuancheng Ren and Yang Fan and Yang Su and Yichang Zhang and Yinger Zhang and Yu Wan and Yuqiong Liu and Zekun Wang and Zeyu Cui and Zhenru Zhang and Zhipeng Zhou and Zihan Qiu},
    journal = {arXiv preprint arXiv:2505.09388},
    year={2025}
}

@inproceedings{feng2024tasl,
  title={TaSL: Continual Dialog State Tracking via Task Skill Localization and Consolidation},
  author={Feng, Yujie and Chu, Xu and Xu, Yongxin and Shi, Guangyuan and Liu, Bo and Wu, Xiao-Ming},
  booktitle={Proceedings of the 62nd Annual Meeting of the Association for Computational Linguistics (Volume 1: Long Papers)},
  pages={1266--1279},
  year={2024}
}

@article{razdaibiedina2023progressive,
  title={Progressive prompts: Continual learning for language models},
  author={Razdaibiedina, Anastasia and Mao, Yuning and Hou, Rui and Khabsa, Madian and Lewis, Mike and Almahairi, Amjad},
  journal={arXiv preprint arXiv:2301.12314},
  year={2023}
}

@inproceedings{wang2023orthogonal,
  title={Orthogonal Subspace Learning for Language Model Continual Learning},
  author={Wang, Xiao and Chen, Tianze and Ge, Qiming and Xia, Han and Bao, Rong and Zheng, Rui and Zhang, Qi and Gui, Tao and Huang, Xuan-Jing},
  booktitle={Findings of the Association for Computational Linguistics: EMNLP 2023},
  pages={10658--10671},
  year={2023}
}

@article{wang2022super,
  title={Super-naturalinstructions: Generalization via declarative instructions on 1600+ nlp tasks},
  author={Wang, Yizhong and Mishra, Swaroop and Alipoormolabashi, Pegah and Kordi, Yeganeh and Mirzaei, Amirreza and Arunkumar, Anjana and Ashok, Arjun and Dhanasekaran, Arut Selvan and Naik, Atharva and Stap, David and others},
  journal={arXiv preprint arXiv:2204.07705},
  year={2022}
}

@inproceedings{du2024unlocking,
  title={Unlocking Continual Learning Abilities in Language Models},
  author={Du, Wenyu and Cheng, Shuang and Luo, Tongxu and Qiu, Zihan and Huang, Zeyu and Cheung, Ka Chun and Cheng, Reynold and Fu, Jie},
  booktitle={Findings of the Association for Computational Linguistics: EMNLP 2024},
  pages={6503--6522},
  year={2024}
}

@article{feng2025recurrent,
  title={Recurrent knowledge identification and fusion for language model continual learning},
  author={Feng, Yujie and Wang, Xujia and Lu, Zexin and Fu, Shenghong and Shi, Guangyuan and Xu, Yongxin and Wang, Yasha and Yu, Philip S and Chu, Xu and Wu, Xiao-Ming},
  journal={arXiv preprint arXiv:2502.17510},
  year={2025}
}

@inproceedings{dou-etal-2024-loramoe,
    title = "{L}o{RAM}o{E}: Alleviating World Knowledge Forgetting in Large Language Models via {M}o{E}-Style Plugin",
    author = "Dou, Shihan  and
      Zhou, Enyu  and
      Liu, Yan  and
      Gao, Songyang  and
      Shen, Wei  and
      Xiong, Limao  and
      Zhou, Yuhao  and
      Wang, Xiao  and
      Xi, Zhiheng  and
      Fan, Xiaoran  and
      Pu, Shiliang  and
      Zhu, Jiang  and
      Zheng, Rui  and
      Gui, Tao  and
      Zhang, Qi  and
      Huang, Xuanjing",
    editor = "Ku, Lun-Wei  and
      Martins, Andre  and
      Srikumar, Vivek",
    booktitle = "Proceedings of the 62nd Annual Meeting of the Association for Computational Linguistics (Volume 1: Long Papers)",
    month = aug,
    year = "2024",
    address = "Bangkok, Thailand",
    publisher = "Association for Computational Linguistics",
    url = "https://aclanthology.org/2024.acl-long.106",
    doi = "10.18653/v1/2024.acl-long.106",
    pages = "1932--1945",

}

@article{eskandar2025star,
  title={STAR: Stability-Inducing Weight Perturbation for Continual Learning},
  author={Eskandar, Masih and Imtiaz, Tooba and Hill, Davin and Wang, Zifeng and Dy, Jennifer},
  journal={arXiv preprint arXiv:2503.01595},
  year={2025}
}

@article{ke2021achieving,
  title={Achieving forgetting prevention and knowledge transfer in continual learning},
  author={Ke, Zixuan and Liu, Bing and Ma, Nianzu and Xu, Hu and Shu, Lei},
  journal={Advances in Neural Information Processing Systems},
  volume={34},
  pages={22443--22456},
  year={2021}
}

@article{zhang2015character,
  title={Character-level convolutional networks for text classification},
  author={Zhang, Xiang and Zhao, Junbo and LeCun, Yann},
  journal={Advances in neural information processing systems},
  volume={28},
  year={2015}
}

@inproceedings{razdaibiedina2022progressive,
  title={Progressive Prompts: Continual Learning for Language Models},
  author={Razdai, Anastasia and Mao, Yuning and Hou, Rui and Khabsa, Madian and Lewis, Mike and Almahairi, Amjad},
  booktitle={The Eleventh International Conference on Learning Representations},
  year={2022}
}

@incollection{mccloskey1989catastrophic,
  title={Catastrophic interference in connectionist networks: The sequential learning problem},
  author={McCloskey, Michael and Cohen, Neal J},
  booktitle={Psychology of learning and motivation},
  volume={24},
  pages={109--165},
  year={1989},
  publisher={Elsevier}
}

@article{chang2024survey,
  title={A survey on evaluation of large language models},
  author={Chang, Yupeng and Wang, Xu and Wang, Jindong and Wu, Yuan and Yang, Linyi and Zhu, Kaijie and Chen, Hao and Yi, Xiaoyuan and Wang, Cunxiang and Wang, Yidong and others},
  journal={ACM Transactions on Intelligent Systems and Technology},
  volume={15},
  number={3},
  pages={1--45},
  year={2024},
  publisher={ACM New York, NY}
}

@article{yu2024recent,
  title={Recent Advances of Multimodal Continual Learning: A Comprehensive Survey},
  author={Yu, Dianzhi and Zhang, Xinni and Chen, Yankai and Liu, Aiwei and Zhang, Yifei and Yu, Philip S and King, Irwin},
  journal={arXiv preprint arXiv:2410.05352},
  year={2024}
}

@inproceedings{chumodel,
  title={MODEL SHAPLEY: Find Your Ideal Parameter Player via One Gradient Backpropagation},
  author={Chu, Xu and Jiang, Xinke and Qiu, Rihong and Gao, Jiaran and Zhao, Junfeng},
  booktitle={The Thirty-ninth Annual Conference on Neural Information Processing Systems},
  year={2025}
}

@inproceedings{zhou2025dropping,
  title={Dropping Experts, Recombining Neurons: Retraining-Free Pruning for Sparse Mixture-of-Experts LLMs},
  author={Zhou, Yixiao and Zhao, Ziyu and Cheng, Dongzhou and Wu, Zhiliang and Gui, Jie and Yang, Yi and Wu, Fei and Cheng, Yu and Fan, Hehe},
  booktitle={Findings of the Association for Computational Linguistics: EMNLP 2025},
  pages={15169--15186},
  year={2025}
}

@inproceedings{hu-etal-2025-longrecipe,
    title = "{L}ong{R}ecipe: Recipe for Efficient Long Context Generalization in Large Language Models",
    author = "Hu, Zhiyuan  and
      Liu, Yuliang  and
      Zhao, Jinman  and
      Wang, Suyuchen  and
      WangYan, WangYan  and
      Shen, Wei  and
      Gu, Qing  and
      Luu, Anh Tuan  and
      Ng, See-Kiong  and
      Jiang, Zhiwei  and
      Hooi, Bryan",
    editor = "Che, Wanxiang  and
      Nabende, Joyce  and
      Shutova, Ekaterina  and
      Pilehvar, Mohammad Taher",
    booktitle = "Proceedings of the 63rd Annual Meeting of the Association for Computational Linguistics (Volume 1: Long Papers)",
    month = jul,
    year = "2025",
    address = "Vienna, Austria",
    publisher = "Association for Computational Linguistics",
    url = "https://aclanthology.org/2025.acl-long.581/",
    pages = "11857--11870",
    ISBN = "979-8-89176-251-0"
}

@misc{kang2025hssbenchbenchmarkinghumanitiessocial,
      title={HSSBench: Benchmarking Humanities and Social Sciences Ability for Multimodal Large Language Models}, 
      author={Zhaolu Kang and Junhao Gong and Jiaxu Yan and Wanke Xia and Yian Wang and Ziwen Wang and Huaxuan Ding and Zhuo Cheng and Wenhao Cao and Zhiyuan Feng and Siqi He and Shannan Yan and Junzhe Chen and Xiaomin He and Chaoya Jiang and Wei Ye and Kaidong Yu and Xuelong Li},
      year={2025},
      eprint={2506.03922},
      archivePrefix={arXiv},
      primaryClass={cs.CL},
      url={https://arxiv.org/abs/2506.03922}, 
}

@inproceedings{
zhao2024large,
title={Large Language Model is not a (Multilingual) Compositional Relation Reasoner},
author={Jinman Zhao and Xueyan Zhang},
booktitle={First Conference on Language Modeling},
year={2024},
url={https://openreview.net/forum?id=wLQ3I0F1oj}
}

@inproceedings{
chen2024entity,
title={Entity Alignment with Noisy Annotations from Large Language Models},
author={Shengyuan Chen and Qinggang Zhang and Junnan Dong and Wen Hua and Qing Li and Xiao Huang},
booktitle={The Thirty-eighth Annual Conference on Neural Information Processing Systems},
year={2024},
}

@inproceedings{logicrag,
title={You Don't Need Pre-built Graphs for {RAG}: Retrieval Augmented Generation with Adaptive Reasoning Structures},
author={Shengyuan Chen and Chuang Zhou and Zheng Yuan and Qinggang Zhang and Zeyang Cui and Hao Chen and Yilin
Xiao and Jiannong Cao and Xiao Huang},
booktitle={The Fortieth AAAI Conference on Artificial Intelligence},
year={2025}
}

@inproceedings{xu2025parenting,
  title={Parenting: Optimizing knowledge selection of retrieval-augmented language models with parameter decoupling and tailored tuning},
  author={Xu, Yongxin and Zhang, Ruizhe and Jiang, Xinke and Feng, Yujie and Xiao, Yuzhen and Ma, Xinyu and Zhu, Runchuan and Chu, Xu and Zhao, Junfeng and Wang, Yasha},
  booktitle={Proceedings of the 63rd Annual Meeting of the Association for Computational Linguistics (Volume 1: Long Papers)},
  pages={11643--11662},
  year={2025}
}

@article{shi2024understanding,
  title={Understanding layer significance in llm alignment},
  author={Shi, Guangyuan and Lu, Zexin and Dong, Xiaoyu and Zhang, Wenlong and Zhang, Xuanyu and Feng, Yujie and Wu, Xiao-Ming},
  journal={arXiv preprint arXiv:2410.17875},
  year={2024}
}

@inproceedings{feng2024continual,
  title={Continual dialogue state tracking via reason-of-select distillation},
  author={Feng, Yujie and Liu, Bo and Dong, Xiaoyu and Lu, Zexin and Zhan, Li-Ming and Wu, Xiao-Ming and Lam, Albert},
  booktitle={Findings of the Association for Computational Linguistics: ACL 2024},
  pages={7075--7087},
  year={2024}
}

@inproceedings{dong2024zero,
  title={Zero-shot Cross-domain Dialogue State Tracking via Context-aware Auto-prompting and Instruction-following Contrastive Decoding},
  author={Dong, Xiaoyu and Feng, Yujie and Lu, Zexin and Shi, Guangyuan and Wu, Xiao-Ming},
  booktitle={Proceedings of the 2024 Conference on Empirical Methods in Natural Language Processing},
  pages={8527--8540},
  year={2024}
}

@article{feng2023towards,
  title={Towards LLM-driven dialogue state tracking},
  author={Feng, Yujie and Lu, Zexin and Liu, Bo and Zhan, Liming and Wu, Xiao-Ming},
  journal={arXiv preprint arXiv:2310.14970},
  year={2023}
}

@misc{hu2021loralowrankadaptationlarge,
      title={LoRA: Low-Rank Adaptation of Large Language Models}, 
      author={Edward J. Hu and Yelong Shen and Phillip Wallis and Zeyuan Allen-Zhu and Yuanzhi Li and Shean Wang and Lu Wang and Weizhu Chen},
      year={2021},
      eprint={2106.09685},
      archivePrefix={arXiv},
      primaryClass={cs.CL},
      url={https://arxiv.org/abs/2106.09685}, 
}

@inproceedings{chustackelberg,
  title={Stackelberg Self-Annotation: A Robust Approach to Data-Efficient LLM Alignment},
  author={Chu, Xu and Zhang, Zhixin and Jia, Tianyu and Jin, Yujie},
  booktitle={The Thirty-ninth Annual Conference on Neural Information Processing Systems},
  year={2025}
}

@inproceedings{wang2025recall,
  title={RECALL: REpresentation-aligned Catastrophic-forgetting ALLeviation via Hierarchical Model Merging},
  author={Wang, Bowen and Wan, Haiyuan and Shi, Liwen and Yang, Chen and He, Peng and Ma, Yue and Han, Haochen and Li, Wenhao and Tan, Tiao and Li, Yongjian and others},
  booktitle={Proceedings of the 2025 Conference on Empirical Methods in Natural Language Processing},
  pages={16392--16406},
  year={2025}
}

@misc{wu2025humanmemoryaimemory,
      title={From Human Memory to AI Memory: A Survey on Memory Mechanisms in the Era of LLMs}, 
      author={Yaxiong Wu and Sheng Liang and Chen Zhang and Yichao Wang and Yongyue Zhang and Huifeng Guo and Ruiming Tang and Yong Liu},
      year={2025},
      eprint={2504.15965},
      archivePrefix={arXiv},
      primaryClass={cs.IR},
      url={https://arxiv.org/abs/2504.15965}, 
}

@inproceedings{feng2025aimmerging,
  title={AIMMerging: Adaptive Iterative Model Merging Using Training Trajectories for Language Model Continual Learning},
  author={Feng, Yujie and Li, Jian and Dong, Xiaoyu and Xu, Pengfei and Zhou, Xiaohui and Zhang, Yujia and Lu, Zexin and Wang, Yasha and Zhao, Alan and Chu, Xu and others},
  booktitle={Proceedings of the 2025 Conference on Empirical Methods in Natural Language Processing},
  pages={13431--13448},
  year={2025}
}

@inproceedings{zeng-etal-2025-task,
    title = "Task-wrapped Continual Learning in Task-Oriented Dialogue Systems",
    author = "Zeng, Min  and
      Yang, Haiqin  and
      Chen, Xi  and
      Guo, Yike",
    editor = "Chiruzzo, Luis  and
      Ritter, Alan  and
      Wang, Lu",
    booktitle = "Findings of the Association for Computational Linguistics: NAACL 2025",
    month = apr,
    year = "2025",
    address = "Albuquerque, New Mexico",
    publisher = "Association for Computational Linguistics",
    url = "https://aclanthology.org/2025.findings-naacl.174/",
    doi = "10.18653/v1/2025.findings-naacl.174",
    pages = "3173--3183",
    ISBN = "979-8-89176-195-7"
}

@misc{bai2025efficientrehearsalschemecatastrophic,
      title={An Efficient Rehearsal Scheme for Catastrophic Forgetting Mitigation during Multi-stage Fine-tuning}, 
      author={Andrew Bai and Chih-Kuan Yeh and Cho-Jui Hsieh and Ankur Taly},
      year={2025},
      eprint={2402.08096},
      archivePrefix={arXiv},
      primaryClass={cs.LG},
      url={https://arxiv.org/abs/2402.08096}, 
}

@inproceedings{ong2025towards,
  title={Towards lifelong dialogue agents via timeline-based memory management},
  author={Ong, Kai Tzu-iunn and Kim, Namyoung and Gwak, Minju and Chae, Hyungjoo and Kwon, Taeyoon and Jo, Yohan and Hwang, Seung-won and Lee, Dongha and Yeo, Jinyoung},
  booktitle={Proceedings of the 2025 Conference of the Nations of the Americas Chapter of the Association for Computational Linguistics: Human Language Technologies (Volume 1: Long Papers)},
  pages={8631--8661},
  year={2025}
}

@inproceedings{chen2025prototype,
  title={Prototype Conditioned Generative Replay for Continual Learning in NLP},
  author={Chen, Xi and Zeng, Min},
  booktitle={Proceedings of the 2025 Conference of the Nations of the Americas Chapter of the Association for Computational Linguistics: Human Language Technologies (Volume 1: Long Papers)},
  pages={12754--12770},
  year={2025}
}

@article{murre2015replication,
  title={Replication and analysis of Ebbinghaus’ forgetting curve},
  author={Murre, Jaap MJ and Dros, Joeri},
  journal={PloS one},
  volume={10},
  number={7},
  pages={e0120644},
  year={2015},
  publisher={Public Library of Science San Francisco, CA USA}
}

@article{wu2025human,
  title={From human memory to ai memory: A survey on memory mechanisms in the era of llms},
  author={Wu, Yaxiong and Liang, Sheng and Zhang, Chen and Wang, Yichao and Zhang, Yongyue and Guo, Huifeng and Tang, Ruiming and Liu, Yong},
  journal={arXiv preprint arXiv:2504.15965},
  year={2025}
}

@article{zhang2025survey,
  title={A survey on the memory mechanism of large language model-based agents},
  author={Zhang, Zeyu and Dai, Quanyu and Bo, Xiaohe and Ma, Chen and Li, Rui and Chen, Xu and Zhu, Jieming and Dong, Zhenhua and Wen, Ji-Rong},
  journal={ACM Transactions on Information Systems},
  volume={43},
  number={6},
  pages={1--47},
  year={2025},
  publisher={ACM New York, NY}
}

@article{liu2025advances,
  title={Advances and challenges in foundation agents: From brain-inspired intelligence to evolutionary, collaborative, and safe systems},
  author={Liu, Bang and Li, Xinfeng and Zhang, Jiayi and Wang, Jinlin and He, Tanjin and Hong, Sirui and Liu, Hongzhang and Zhang, Shaokun and Song, Kaitao and Zhu, Kunlun and others},
  journal={arXiv preprint arXiv:2504.01990},
  year={2025}
}

@InProceedings{pmlr-v267-wan25d,
  title = 	 {Understanding the Forgetting of ({R}eplay-based) Continual Learning via Feature Learning: Angle Matters},
  author =       {Wan, Hongyi and Ren, Shiyuan and Huang, Wei and Zhang, Miao and Deng, Xiang and Bao, Yixin and Nie, Liqiang},
  booktitle = 	 {Proceedings of the 42nd International Conference on Machine Learning},
  pages = 	 {61956--62019},
  year = 	 {2025},
  editor = 	 {Singh, Aarti and Fazel, Maryam and Hsu, Daniel and Lacoste-Julien, Simon and Berkenkamp, Felix and Maharaj, Tegan and Wagstaff, Kiri and Zhu, Jerry},
  volume = 	 {267},
  series = 	 {Proceedings of Machine Learning Research},
  month = 	 {13--19 Jul},
  publisher =    {PMLR},
  url = 	 {https://proceedings.mlr.press/v267/wan25d.html},
}

@misc{lu2025rethinkingstabilityplasticitytradeoffcontinual,
      title={Rethinking the Stability-Plasticity Trade-off in Continual Learning from an Architectural Perspective}, 
      author={Aojun Lu and Hangjie Yuan and Tao Feng and Yanan Sun},
      year={2025},
      eprint={2506.03951},
      archivePrefix={arXiv},
      primaryClass={cs.LG},
      url={https://arxiv.org/abs/2506.03951}, 
}

@misc{deng2025unlockingpowerrehearsalcontinual,
      title={Unlocking the Power of Rehearsal in Continual Learning: A Theoretical Perspective}, 
      author={Junze Deng and Qinhang Wu and Peizhong Ju and Sen Lin and Yingbin Liang and Ness Shroff},
      year={2025},
      eprint={2506.00205},
      archivePrefix={arXiv},
      primaryClass={cs.LG},
      url={https://arxiv.org/abs/2506.00205}, 
}

@misc{tang2025mergingmodelsflyretraining,
      title={Merging Models on the Fly Without Retraining: A Sequential Approach to Scalable Continual Model Merging}, 
      author={Anke Tang and Enneng Yang and Li Shen and Yong Luo and Han Hu and Bo Du and Dacheng Tao},
      year={2025},
      eprint={2501.09522},
      archivePrefix={arXiv},
      primaryClass={cs.LG},
      url={https://arxiv.org/abs/2501.09522}, 
}

@article{momeni2025anacp,
  title={AnaCP: Toward Upper-Bound Continual Learning via Analytic Contrastive Projection},
  author={Momeni, Saleh and Xiao, Changnan and Liu, Bing},
  journal={arXiv preprint arXiv:2511.13880},
  year={2025}
}

@article{cheng2025continuous,
  title={Continuous Subspace Optimization for Continual Learning},
  author={Cheng, Quan and Wan, Yuanyu and Wu, Lingyu and Hou, Chenping and Zhang, Lijun},
  journal={arXiv preprint arXiv:2505.11816},
  year={2025}
}

@inproceedings{huang2025mitigating,
  title={Mitigating Catastrophic Forgetting in Large Language Models with Forgetting-aware Pruning},
  author={Huang, Wei and Cheng, Anda and Wang, Yinggui},
  booktitle={Proceedings of the 2025 Conference on Empirical Methods in Natural Language Processing},
  pages={21853--21867},
  year={2025}
}

@inproceedings{li2025dynamic,
  title={Dynamic Expert Specialization: Towards Catastrophic Forgetting-Free Multi-Domain MoE Adaptation},
  author={Li, Junzhuo and Wang, Bo and Zhou, Xiuze and Hu, Xuming},
  booktitle={Proceedings of the 2025 Conference on Empirical Methods in Natural Language Processing},
  pages={18489--18504},
  year={2025}
}

@inproceedings{wan2025empowering,
  title={Empowering Math Problem Generation and Reasoning for Large Language Model via Synthetic Data based Continual Learning Framework},
  author={Wan, Qian and Shi, Wangzi and Feng, Jintian and Liu, Shengyingjie and Wei, Luona and Dai, Zhicheng and Sun, Jianwen},
  booktitle={Proceedings of the 2025 Conference on Empirical Methods in Natural Language Processing},
  pages={23983--24002},
  year={2025}
}

@inproceedings{kang2025your,
  title={Do Your Best and Get Enough Rest for Continual Learning},
  author={Kang, Hankyul and Seifer, Gregor and Lee, Donghyun and Ryu, Jongbin},
  booktitle={Proceedings of the Computer Vision and Pattern Recognition Conference},
  pages={10077--10086},
  year={2025}
}

@article{chen2025incorporating,
  title={Incorporating Forgetting Curve and Memory Replay for Evolving Socially-aware Recommendation},
  author={Chen, Hongqi and Feng, Zhiyong and Chen, Shizhan and Wu, Hongyue and Sun, Yingchao and Li, Jingyu and Gao, Qinghang and Zhang, Lu and Xue, Xiao},
  journal={Information Processing \& Management},
  volume={62},
  number={3},
  pages={104070},
  year={2025},
  publisher={Elsevier}
}

@article{zhang2025merge,
  title={Merge then Realign: Simple and Effective Modality-Incremental Continual Learning for Multimodal LLMs},
  author={Zhang, Dingkun and Qi, Shuhan and Xiao, Xinyu and Chen, Kehai and Wang, Xuan},
  journal={arXiv preprint arXiv:2503.07663},
  year={2025}
}

@inproceedings{zeng2025modalprompt,
  title={Modalprompt: Towards efficient multimodal continual instruction tuning with dual-modality guided prompt},
  author={Zeng, Fanhu and Zhu, Fei and Guo, Haiyang and Zhang, Xu-Yao and Liu, Cheng-Lin},
  booktitle={Proceedings of the 2025 Conference on Empirical Methods in Natural Language Processing},
  pages={12137--12152},
  year={2025}
}

@article{jiang2024interpretable,
  title={Interpretable catastrophic forgetting of large language model fine-tuning via instruction vector},
  author={Jiang, Gangwei and Jiang, Caigao and Li, Zhaoyi and Xue, Siqiao and Zhou, Jun and Song, Linqi and Lian, Defu and Wei, Ying},
  journal={arXiv preprint arXiv:2406.12227},
  year={2024}
}

@article{feng2024tasl2,
  title={KIF: Knowledge Identification and Fusion for Language Model Continual Learning},
  author={Feng, Yujie and Chu, Xu and Xu, Yongxin and Lu, Zexin and Liu, Bo and Yu, Philip S and Wu, Xiao-Ming},
  journal={arXiv preprint arXiv:2408.05200},
  year={2024}
}

@article{ren2024analyzing,
  title={Analyzing and Reducing Catastrophic Forgetting in Parameter Efficient Tuning},
  author={Ren, Weijieying and Li, Xinlong and Wang, Lei and Zhao, Tianxiang and Qin, Wei},
  journal={arXiv preprint arXiv:2402.18865},
  year={2024}
}

@inproceedings{alexandrov2024mitigating,
  title={Mitigating Catastrophic Forgetting in Language Transfer via Model Merging},
  author={Alexandrov, Anton and Raychev, Veselin and Mueller, Mark and Zhang, Ce and Vechev, Martin and Toutanova, Kristina},
  booktitle={Findings of the Association for Computational Linguistics: EMNLP 2024},
  pages={17167--17186},
  year={2024}
}

@article{jiang2025unlocking,
  title={Unlocking the power of function vectors for characterizing and mitigating catastrophic forgetting in continual instruction tuning},
  author={Jiang, Gangwei and Jiang, Caigao and Li, Zhaoyi and Xue, Siqiao and Zhou, Jun and Song, Linqi and Lian, Defu and Wei, Ying},
  journal={arXiv preprint arXiv:2502.11019},
  year={2025}
}

@article{wang2024self,
  title={Self-Expansion of Pre-trained Models with Mixture of Adapters for Continual Learning},
  author={Wang, Huiyi and Lu, Haodong and Yao, Lina and Gong, Dong},
  journal={arXiv preprint arXiv:2403.18886},
  year={2024}
}

@article{he2024seekr,
  title={SEEKR: Selective Attention-Guided Knowledge Retention for Continual Learning of Large Language Models},
  author={He, Jinghan and Guo, Haiyun and Zhu, Kuan and Zhao, Zihan and Tang, Ming and Wang, Jinqiao},
  journal={arXiv preprint arXiv:2411.06171},
  year={2024}
}

@article{li2024revisiting,
  title={Revisiting Catastrophic Forgetting in Large Language Model Tuning},
  author={Li, Hongyu and Ding, Liang and Fang, Meng and Tao, Dacheng},
  journal={arXiv preprint arXiv:2406.04836},
  year={2024}
}

@article{ke2023sub,
  title={Sub-network Discovery and Soft-masking for Continual Learning of Mixed Tasks},
  author={Ke, Zixuan and Liu, Bing and Xiong, Wenhan and Celikyilmaz, Asli and Li, Haoran},
  journal={arXiv preprint arXiv:2310.09436},
  year={2023}
}

@article{kline2025human,
  title={Human-like Forgetting Curves in Deep Neural Networks},
  author={Kline, Dylan},
  journal={arXiv preprint arXiv:2506.12034},
  year={2025}
}

@article{pham2023continual,
  title={Continual learning, fast and slow},
  author={Pham, Quang and Liu, Chenghao and Hoi, Steven CH},
  journal={IEEE Transactions on Pattern Analysis and Machine Intelligence},
  year={2023},
  publisher={IEEE}
}

@article{wang2025see,
  title={SEE: Continual Fine-tuning with Sequential Ensemble of Experts},
  author={Wang, Zhilin and Li, Yafu and Qu, Xiaoye and Cheng, Yu},
  journal={arXiv preprint arXiv:2504.06664},
  year={2025}
}

@article{huang2024mitigating,
  title={Mitigating catastrophic forgetting in large language models with self-synthesized rehearsal},
  author={Huang, Jianheng and Cui, Leyang and Wang, Ante and Yang, Chengyi and Liao, Xinting and Song, Linfeng and Yao, Junfeng and Su, Jinsong},
  journal={arXiv preprint arXiv:2403.01244},
  year={2024}
}

@article{zhou2024continual,
  title={Continual learning with pre-trained models: A survey},
  author={Zhou, Da-Wei and Sun, Hai-Long and Ning, Jingyi and Ye, Han-Jia and Zhan, De-Chuan},
  journal={arXiv preprint arXiv:2401.16386},
  year={2024}
}

@inproceedings{zhong2024memorybank,
  title={Memorybank: Enhancing large language models with long-term memory},
  author={Zhong, Wanjun and Guo, Lianghong and Gao, Qiqi and Ye, He and Wang, Yanlin},
  booktitle={Proceedings of the AAAI Conference on Artificial Intelligence},
  volume={38},
  number={17},
  pages={19724--19731},
  year={2024}
}

\appendix
\label{sec:appendix}

\section{Usage of Language Models}
We used a large language model (LLM) for editorial tasks such as proofreading, grammar correction, and enhancing clarity and readability.

\section{Additional Results}
\label{sec:hyper}
\subsection{Generalization to Full-Parameter Fine-Tuning}
\label{sec:full_ft}

We further evaluate whether our method generalizes beyond parameter-efficient fine-tuning. 
While the main experiments are conducted under LoRA-based adaptation, the proposed model-centric time formulation is not tied to any specific parameterization.

To validate this, we conduct additional experiments under full-parameter fine-tuning on SuperNI (Qwen3-0.6B, task order 7). 
Results are summarized in Table~\ref{tab:full_ft}.

FOREVER achieves the best performance on both OP and BWT. 
Compared with the strongest baseline, it improves OP by +1.1 and reduces forgetting (BWT) by +0.9. 
These results demonstrate that our approach remains effective when all model parameters are updated.

Overall, this suggests that the proposed model-centric time calibration is a general mechanism for continual learning, 
rather than being specific to LoRA-based adaptation.

\begin{table}[t]
\centering
\small
\begin{tabular}{lcc}
\toprule
\textbf{Method} & OP $\uparrow$ & BWT $\uparrow$ \\
\midrule
Recurrent-KIF & 46.0 & -4.1 \\
AIMMerging & 46.4 & -3.0 \\
VBM & 46.4 & -3.1 \\
FOREVER & \textbf{47.5} & \textbf{-2.2} \\
\bottomrule
\end{tabular}
\caption{Results under full-parameter fine-tuning.}
\label{tab:full_ft}
\end{table}

\subsection{Effect of the Memory Size}
We examine the effect of varying memory size on the performance of VBM and {\ouralg}.
Specifically, we adjust the memory size per task $|M|$ to \{2\%, 5\%, 10\%, 50\%\}, and report the results on SuperNI benchmark in Table~\ref{tbl:ablation_memory}.

As expected, increasing the memory size consistently improves performance for both methods, since more historical samples provide stronger supervision for mitigating forgetting.
However, under the same memory budget, {\ouralg} consistently outperforms VBM across all memory sizes.
This advantage stems from {\ouralg}'s dynamics-aware replay strategy, which selectively schedules replay based on model-centric time and adaptively controls replay strength.

Notably, the performance gap between {\ouralg} and MixReplay is more pronounced in low-memory regimes (e.g., $|M|=2\%$ and $5\%$), indicating that {\ouralg} makes more efficient use of limited memory resources.
These results suggest that aligning replay with model update dynamics is particularly beneficial when memory capacity is constrained, a setting that is common in practical continual learning scenarios.

\begin{table}[h]
\centering
\scalebox{1}{
\begin{tabular}{lcccc}
\toprule
\multirow{2}*{\tabincell{c}{ }} & \multicolumn{4}{c}{Memory Size}\\
\cmidrule(lr){2-5}
 & 2\% & 5\% & 10\% & 50\%\\
\midrule

\rule{0pt}{6pt} VBM & 41.2 & 42.4& 43.0 & 45.4   \\

\rule{0pt}{8pt} {\ouralg} & 42.5 & 43.5  & 43.9 & 46.4 \\

\bottomrule
\end{tabular}}
\caption{Ablation study on memory size, using Qwen3-0.6B as the backbone.
}
\label{tbl:ablation_memory}
\end{table}

\subsection{Sensitivity Analysis of Hyperparameters}

We analyze the sensitivity of {\ouralg} to the warm-up length $S$, which is used to calibrate the virtual model day and the baseline update intensity.
Intuitively, $S$ controls the scale of model-centric time but should not critically affect replay behavior as long as it provides a stable estimate of early training dynamics.
We vary $S$ from $\{12, 24, 48\}$ steps, where $S=24$ is the default choice used in all main experiments.
The results on the SuperNI benchmark (task order~4) are reported in Table~\ref{tbl:ablation_S}.

Overall, {\ouralg} exhibits robust performance across different values of $S$.
While extremely small warm-up windows may introduce minor variance due to noisier update estimates, performance remains stable once $S$ is sufficiently large.
Notably, $S=24$ consistently achieves strong results across all metrics, striking a good balance between calibration stability and computational overhead.

These results indicate that {\ouralg} is not sensitive to the precise choice of $S$, confirming that the proposed model-centric time calibration serves as a coarse-grained normalization of learning dynamics rather than a finely tuned hyperparameter.

\begin{table}[h]
\centering
\scalebox{1}{
\begin{tabular}{ccc}
\toprule
Warm-up $S$ & OP ($\uparrow$) & BWT ($\uparrow$) \\
\midrule
12  & 42.0 & -3.4 \\
24  & \textbf{42.5} &  \textbf{-2.8} \\
48  & 42.3 &  -3.0 \\
\bottomrule
\end{tabular}}
\caption{\textbf{Sensitivity analysis of the warm-up length $S$.}
$S$ controls the calibration of virtual model days and baseline update intensity.
Results are reported on the SuperNI benchmark (task order~7).}
\label{tbl:ablation_S}
\end{table}

\subsection{Sensitivity Analysis of Memory Epochs}
We further analyze the sensitivity of {\ouralg} to the number of replay epochs performed at each replay event.
Specifically, when replay is triggered, we vary the number of training epochs on the memory buffer, denoted as $E_{\text{mem}} \in \{1, 2, 4\}$, while keeping all other settings fixed. $E_{\text{mem}}=2$ is the default choice used in all main experiments.

Table~\ref{tbl:ablation_mem_epoch} reports the results.
Overall, {\ouralg} demonstrates stable performance across different choices of $E_{\text{mem}}$.
Increasing the number of memory epochs generally leads to slightly better retention of previous tasks, reflected by modest improvements in OP, but the gains quickly saturate.
In contrast, using too many replay epochs does not yield proportional benefits and may introduce unnecessary computational overhead.

Based on this observation, we set $E_{\text{mem}} = 2$ as the default choice in all experiments, which achieves a good balance between performance and efficiency.
This result indicates that {\ouralg} does not rely on aggressive replay, and that its effectiveness mainly stems from \emph{when} replay is triggered and \emph{how} replay strength is adaptively controlled, rather than from repeatedly revisiting memory samples.

\begin{table}[h]
\centering
\scalebox{1}{
\begin{tabular}{c|cc}
\toprule
$E_{\text{mem}}$ & OP & BWT \\
\midrule
1 & 42.2 & -3.0 \\
2 & 42.5 & -2.8 \\
4 & 42.6 & -2.9 \\
\bottomrule
\end{tabular}}
\caption{Sensitivity to the number of replay epochs per replay event on the SuperNI benchmark.}
\label{tbl:ablation_mem_epoch}
\end{table}

\subsection{Time Complexity Analysis}
\label{sec:time}

We analyze the computational overhead of {\ouralg} by comparing its training time with representative baseline methods across different model scales, as reported in Table~\ref{tbl:time}.

Overall, {\ouralg} incurs only a modest increase in training time compared to standard replay-based baselines such as MixReplay, while remaining consistently more efficient than step-based Ebbinghaus replay methods (VBM). This efficiency arises from the design of {\ouralg}, which relies solely on lightweight parameter update statistics—specifically, the update magnitude and its exponential moving average—to guide both replay timing and replay strength.

In contrast to model merging–based approaches such as AIMMerging, which require additional bookkeeping or repeated parameter fusion, {\ouralg} avoids expensive auxiliary optimization and task-wise parameter importance estimation. As a result, its training time scales smoothly with model size, making it well suited for large-scale LLM continual learning settings.

Compared to VBM, which triggers replay at fixed step-based intervals and requires additional computation to optimize recall schedules, {\ouralg} schedules replay based on accumulated model evolution. Consequently, replay events are triggered more selectively, leading to fewer unnecessary replay phases and improved runtime efficiency, particularly for larger models.

\begin{table}[ht]
\centering
\scalebox{0.6}{
\begin{tabular}{lrrrr}
\toprule
Training Time \\ (Min/Epoch) & Qwen3-0.6B & Qwen3-4B & LLaMA3.1-8B & LLaMA2-13B  \\
\midrule

\rule{0pt}{4pt}MixReplay	   &1.3 & 3.5 & 5.6 & 6.8	 \\
\rule{0pt}{8pt}AIMMerging	   &1.5 & 4.9 & 8.0 & 9.9	 \\
\rule{0pt}{8pt}VBM   &1.4 & 3.9  & 7.2 & 9.1 \\
\rule{0pt}{8pt}{\ouralg}   &1.4 & 3.8 & 6.9 & 8.5 \\
\bottomrule
\end{tabular}}
\caption{Training time comparison across backbones.}
\label{tbl:time}
\end{table}

\subsection{Empirical Analysis of Forgetting Dynamics}
\label{sec:forgetting_dynamics}

We further investigate whether the fast-then-slow forgetting pattern holds in our continual learning setting. 
Instead of assuming the applicability of cognitive theories, we directly measure forgetting dynamics during training.

Under a representative configuration (Qwen3-0.6B on SuperNI, task order 7), 
we periodically evaluate the loss on previously learned tasks while training on new tasks, 
without updating their gradients. 
We analyze 10 consecutive tasks (task IDs 6--15) and report the average behavior across runs.

To quantify forgetting speed, we divide the training trajectory into consecutive intervals 
(e.g., 0--60, 60--120 steps). 
For each interval, we compute the loss increase $\Delta \text{Loss}$ 
between the end and the beginning of the interval, 
and normalize it by the number of steps $N_{\text{step}}$. 
This gives the average forgetting rate within each phase.
Results are shown in Table~\ref{tab:forgetting_rate}. 

We observe that forgetting is substantially faster in early stages and gradually slows down over time, 
exhibiting a clear fast-then-slow decay pattern. 
The early-stage forgetting rate is approximately 7$\times$ higher than that in later stages, 
indicating strongly non-linear dynamics.
This empirical observation supports the structural motivation of increasing-spacing replay. 
Importantly, we do not assume biological equivalence between human cognition and model training. 
Instead, our findings suggest that the fast-then-slow decay pattern underlying spaced replay 
also emerges in LLM continual learning, providing empirical justification for our design.

\begin{table}[t]
\centering
\small
\begin{tabular}{ccc}
\toprule
\textbf{Training Phase} & \textbf{Step Range} & \textbf{Avg Forgetting Rate} \\
\midrule
1st & 0--60   & 0.0023 \\
2nd & 60--120 & 0.0018 \\
3rd & 120--180 & 0.0010 \\
4th & 180--240 & 0.0005 \\
5th & 240--300 & 0.0003 \\
\bottomrule
\end{tabular}
\caption{Average forgetting rate ($\Delta$Loss/$N_{\text{step}}$) across training phases.}
\label{tab:forgetting_rate}
\end{table}

\subsection{Effect of Initial Window Size $S$}
\label{sec:window_size}

We further analyze the effect of the initial window size $S$, which determines the scale of model-time in our framework. 
Specifically, $S$ defines one virtual ``model day'' as 
$\tau_{\text{day}} = \sum_{i=1}^{S} \Delta \theta_i$, 
i.e., the accumulated parameter change corresponding to one unit of model-time. 
Thus, $S$ rescales the model-time axis but does not alter the underlying increasing-spacing replay structure.

In the main paper, we reported ablation results for $S \in \{12, 24, 48\}$. 
Here, we extend the range to both smaller and larger values to more comprehensively evaluate robustness.

Results on SuperNI (Qwen3-0.6B, task order 7) are shown in Table~\ref{tab:window_size}. 
We observe that performance remains stable across a broad intermediate range (approximately $6$--$48$), forming a clear plateau. 
Noticeable degradation occurs only at extreme values (e.g., $S=3$ or $S=96$).

This behavior is consistent with the role of $S$. 
When $S$ is too small, a model-day becomes short, leading to overly frequent replay, which increases computational overhead and may over-regularize adaptation. 
Conversely, when $S$ is too large, replay becomes too sparse, allowing more forgetting to accumulate.

Overall, these results indicate that $S$ primarily controls the temporal scale of replay rather than its structural behavior. 
As long as $S$ lies within a reasonable range, performance remains robust, suggesting that it is not a highly sensitive hyperparameter. 
We adopt $S=24$ in the main experiments, as it lies within the stable region and consistently achieves strong performance.

\begin{table}[t]
\centering
\small
\begin{tabular}{ccc}
\toprule
\textbf{Initial Window Size $S$} & OP $\uparrow$ & BWT $\uparrow$ \\
\midrule
3   & 41.8 & -3.8 \\
6   & 42.1 & -3.7 \\
12  & 42.0 & -3.4 \\
24  & \textbf{42.5} & \textbf{-2.8} \\
48  & 42.3 & -3.0 \\
96  & 41.2 & -4.4 \\
\bottomrule
\end{tabular}
\caption{Effect of the initial window size $S$ on performance.}
\label{tab:window_size}
\end{table}

\subsection{Scalability to Longer Task Sequences}
\label{sec:long_sequence}

We further evaluate the scalability of our method under longer continual learning horizons. 
While the main experiments consider up to 15 tasks, real-world scenarios often involve substantially longer task sequences.
To this end, we construct a 30-task sequence by combining the Long Sequence benchmark (15 tasks) and SuperNI (15 tasks), 
followed by random shuffling to form a unified continual learning setting. 
Results on Qwen3-0.6B are reported in Table~\ref{tab:long_sequence}.

As the number of tasks increases, all methods experience performance degradation due to accumulated forgetting. 
Nevertheless, FOREVER consistently achieves the strongest performance, improving OP by approximately +1.2 over the strongest baseline, 
while also obtaining better BWT.

These results demonstrate that our model-centric time calibration scales effectively to longer task sequences 
and provides stronger robustness against cumulative forgetting.

\begin{table}[t]
\centering
\small
\begin{tabular}{lcc}
\toprule
\textbf{Method} & OP $\uparrow$ & BWT $\uparrow$ \\
\midrule
Recurrent-KIF & 24.7 & -15.0 \\
AIMMerging & 26.1 & -11.6 \\
VBM & 26.7 & -12.8 \\
FOREVER & \textbf{27.9} & \textbf{-10.8} \\
\bottomrule
\end{tabular}
\caption{Results on a 30-task continual learning setting.}
\label{tab:long_sequence}
\end{table}

\section{Task-Adaptive Behavior of Model-Time $\tau_{\text{day}}$}
\label{sec:tau_analysis}

We further analyze whether the variability of replay trigger timing across tasks reflects meaningful task-dependent dynamics. 
In our framework, replay is triggered based on the model-time $\tau_{\text{day}}$, which is determined by accumulated parameter updates. 
As a result, different tasks may reach the same replay threshold at different training steps. 
We examine whether this variability is consistent, interpretable, and correlated with task properties.

\paragraph{Stability Across Task Orders.}
We first evaluate whether $\tau_{\text{day}}$ is stable under different task orders. 
We compute $\tau_{\text{day}}$ for all 15 tasks in SuperNI under two distinct task sequences (order 7 and order 8). 
The resulting Pearson correlation is $r = 0.7379$ with $p = 0.0017$, indicating a strong and statistically significant correlation.

This result suggests that $\tau_{\text{day}}$ is largely invariant to task order and primarily reflects task-intrinsic learning dynamics, rather than sequence-dependent effects.

\paragraph{Correlation with Final Forgetting (BWT).}
We next examine the relationship between $\tau_{\text{day}}$ and the final backward transfer (BWT) of each task. 
We observe no significant correlation ($r = 0.2182$, $p = 0.4536$). 

This is expected, as BWT is influenced by global cross-task interference and order-dependent interactions, whereas $\tau_{\text{day}}$ captures local parameter drift during the learning of individual tasks. 
Therefore, $\tau_{\text{day}}$ and final BWT measure fundamentally different aspects of the continual learning process.

\paragraph{Correlation with Task Difficulty.}
Finally, we analyze whether $\tau_{\text{day}}$ correlates with task difficulty. 
We use multi-task learning (MTL) performance as a proxy for task difficulty, where lower performance indicates harder tasks.

We observe a statistically significant negative correlation between $\tau_{\text{day}}$ and MTL performance ($r = -0.5625$, $p = 0.0291$). 
This indicates that harder tasks tend to induce larger accumulated parameter updates, resulting in larger $\tau_{\text{day}}$ values.

Overall, these results suggest that $\tau_{\text{day}}$ captures task-dependent parameter dynamics in a consistent and interpretable manner. 
The variability of replay trigger timing across tasks is therefore not incidental, but reflects adaptive behavior aligned with task difficulty and learning dynamics.

\section{Visualization of Task and Replay Regularization Losses}
\label{appendix:loss}

To better understand the role of the replay regularization coefficient and the design choice of $\beta_{\text{base}}$, we visualize the loss components involved during replay events.
Specifically, for each replay stage, we record (i) the task loss on memory samples at the end of replay, and (ii) the scaled regularization loss applied during replay.

\begin{figure}[t]
  \centering
  \includegraphics[width=1\linewidth]{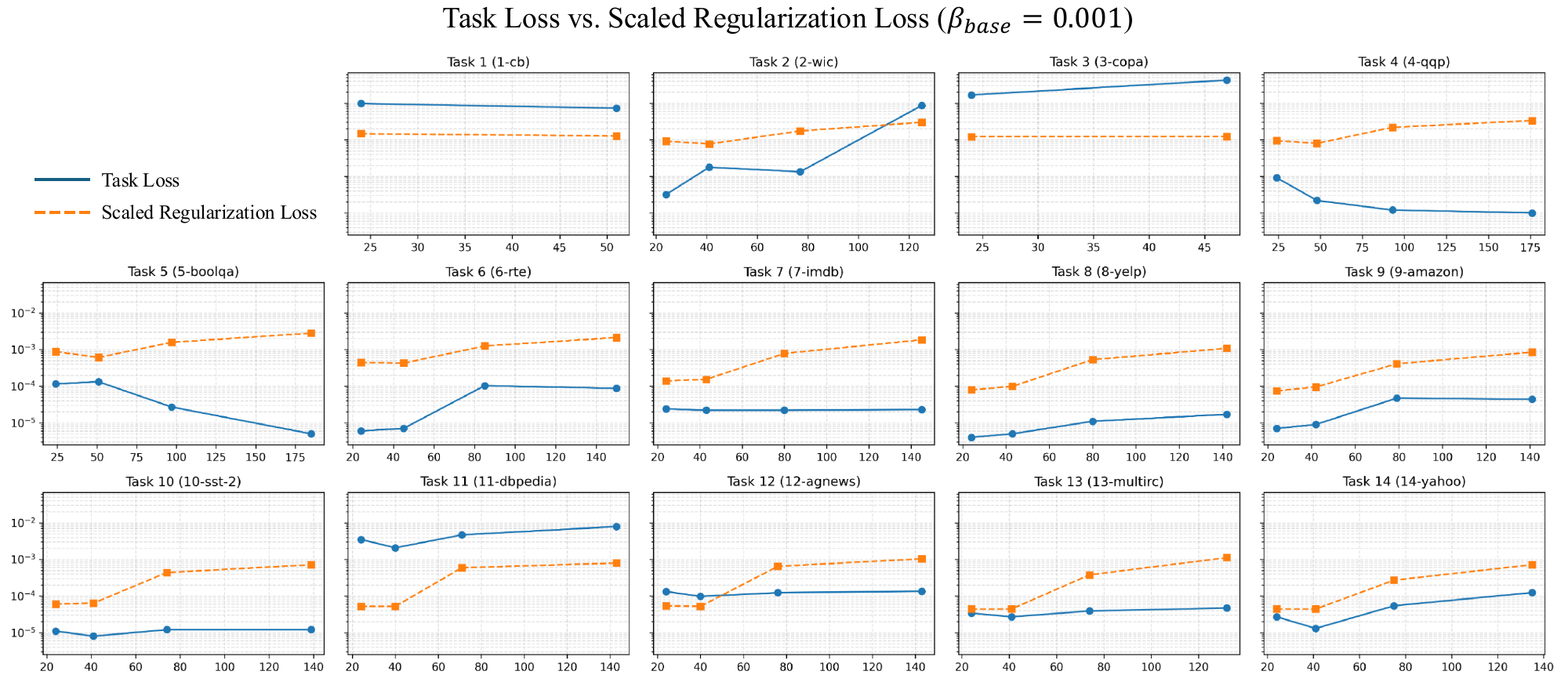}
\caption{Visualization of task loss ($\mathcal{L}_{\text{task}}^{old}$) on memory samples and scaled replay regularization loss ($\beta_{\text{base}} \cdot \mathcal{L}_{\text{reg}}$) at replay stages.}
  \label{fig:regloss}
\end{figure}

Figure~\ref{fig:regloss} shows that after applying the base scaling factor $\beta_{\text{base}}=0.001$, the magnitudes of the task loss and the replay regularization loss are brought onto a comparable numerical scale.
This calibration is critical: without proper scaling, the replay objective would be dominated by either the task loss or the regularization term, preventing effective control of replay strength.

Importantly, this balanced scaling enables the proposed intensity-aware modulation mechanism to operate meaningfully.
Once the two loss components are aligned in magnitude, the adaptive factor derived from update intensity can adjust replay strength without causing instability or overwhelming the task objective.
As a result, replay regularization contributes to stabilizing previously learned knowledge while still allowing sufficient flexibility for new task adaptation.
Overall, this visualization illustrates that $\beta_{\text{base}}$ serves as a necessary normalization factor rather than a tuning heuristic, providing a well-conditioned foundation upon which intensity-aware replay regularization can be effectively applied.

\section{Visualization of Replay Dynamics}
\label{appendix:vis}

\begin{figure*}[t]
    \centering

    \begin{subfigure}[t]{0.48\linewidth}
        \centering
        \includegraphics[width=\linewidth]{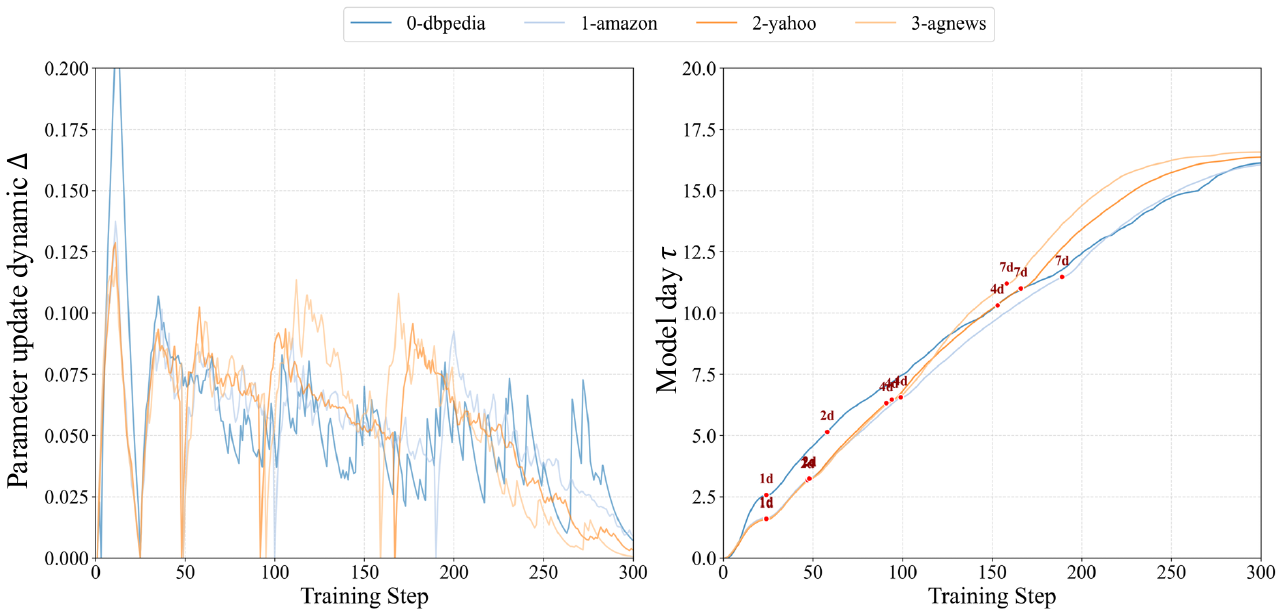}
        \caption{Standard CL Benchmark, task order 1}
    \end{subfigure}
    \hfill
    \begin{subfigure}[t]{0.48\linewidth}
        \centering
        \includegraphics[width=\linewidth]{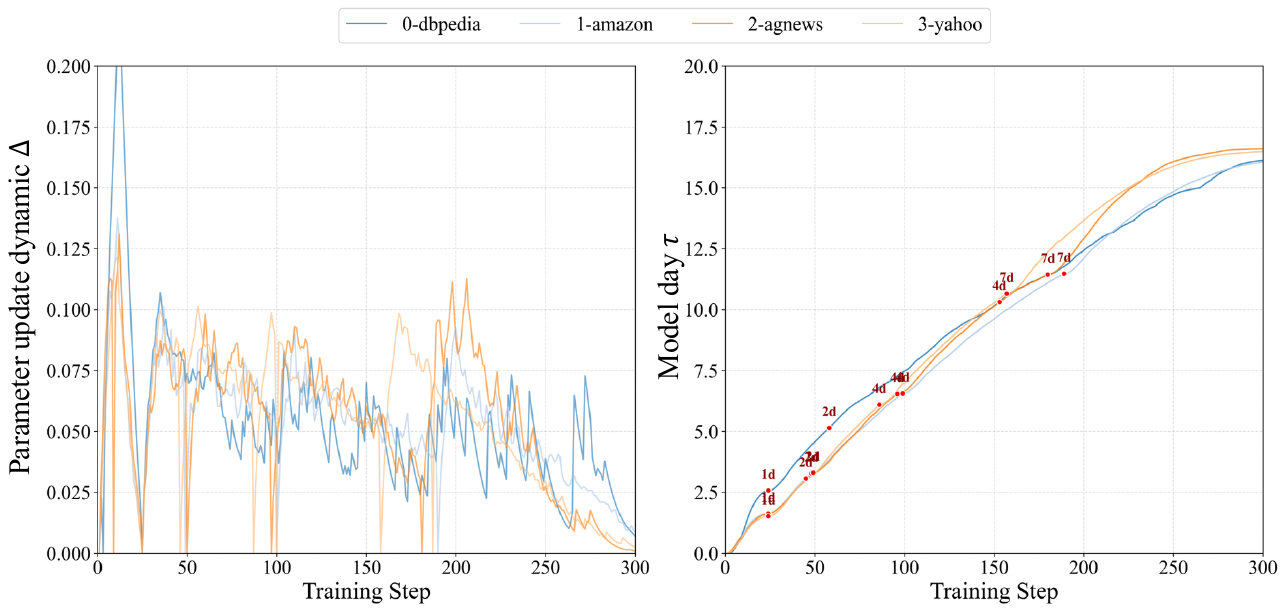}
        \caption{Standard CL Benchmark, task order 2}
    \end{subfigure}

    \vspace{0.4em}

    \begin{subfigure}[t]{0.48\linewidth}
        \centering
        \includegraphics[width=\linewidth]{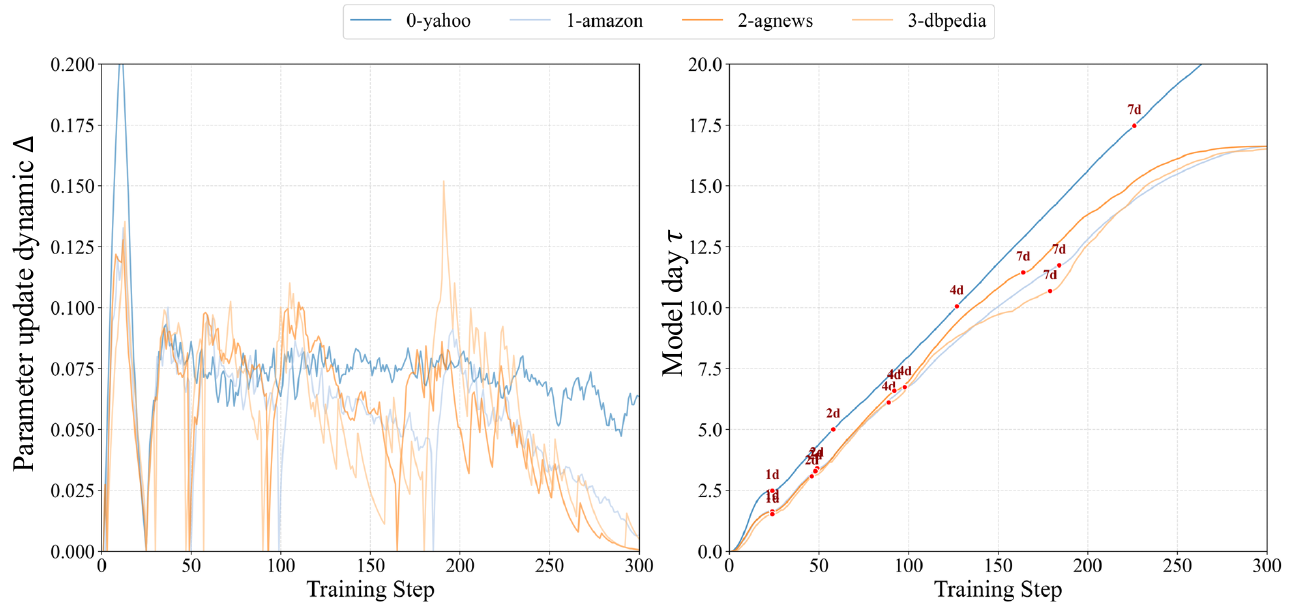}
        \caption{Standard CL Benchmark, task order 3}
    \end{subfigure}
    \hfill
    \begin{subfigure}[t]{0.48\linewidth}
        \centering
        \includegraphics[width=\linewidth]{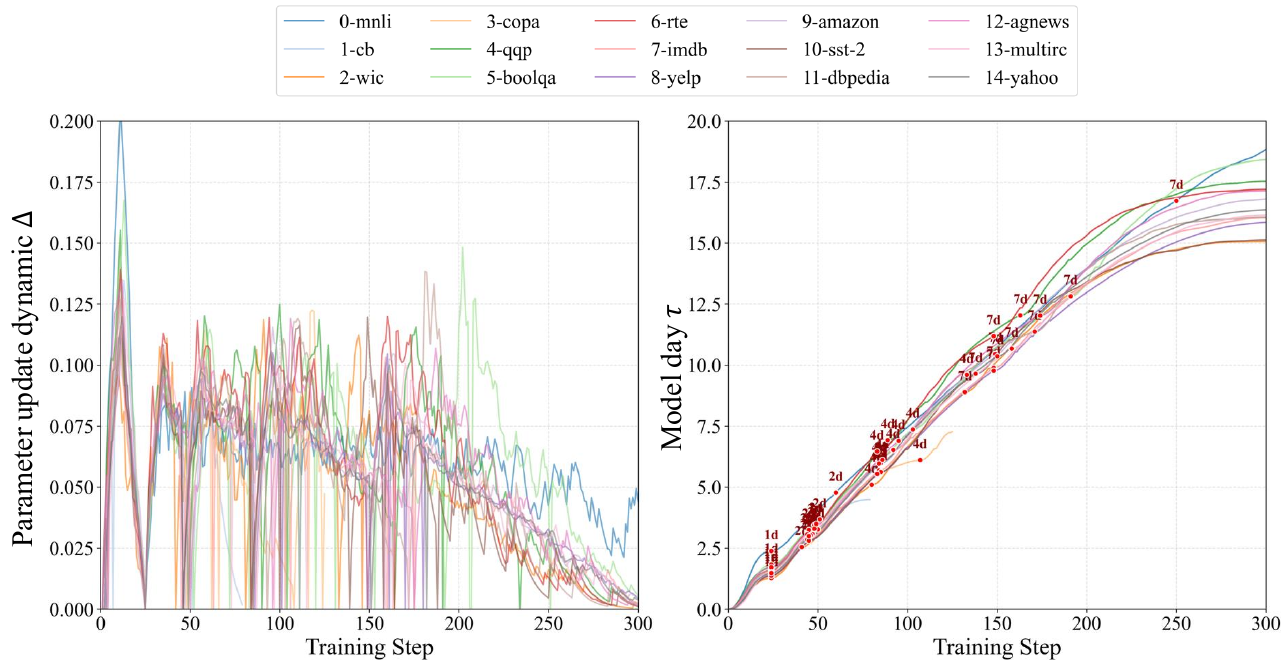}
        \caption{Long Sequence Benchmark, task order 4}
    \end{subfigure}

    \vspace{0.4em}

    \begin{subfigure}[t]{0.48\linewidth}
        \centering
        \includegraphics[width=\linewidth]{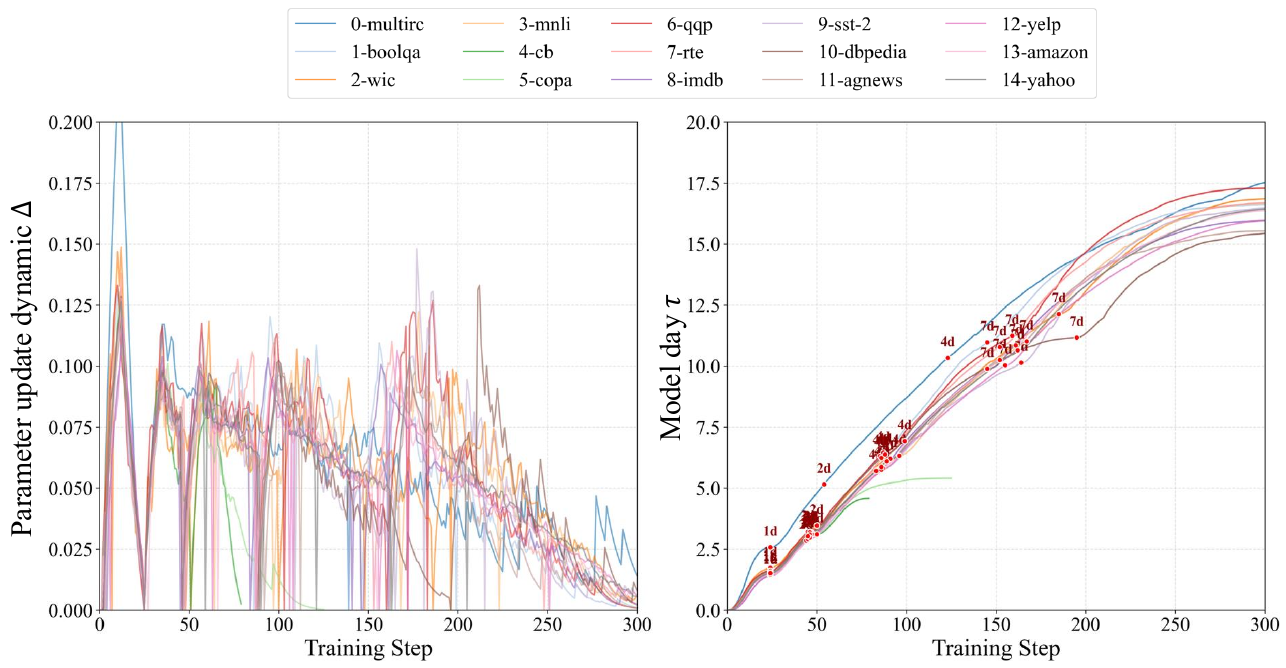}
        \caption{Long Sequence Benchmark, task order 5}
    \end{subfigure}
    \hfill
    \begin{subfigure}[t]{0.48\linewidth}
        \centering
        \includegraphics[width=\linewidth]{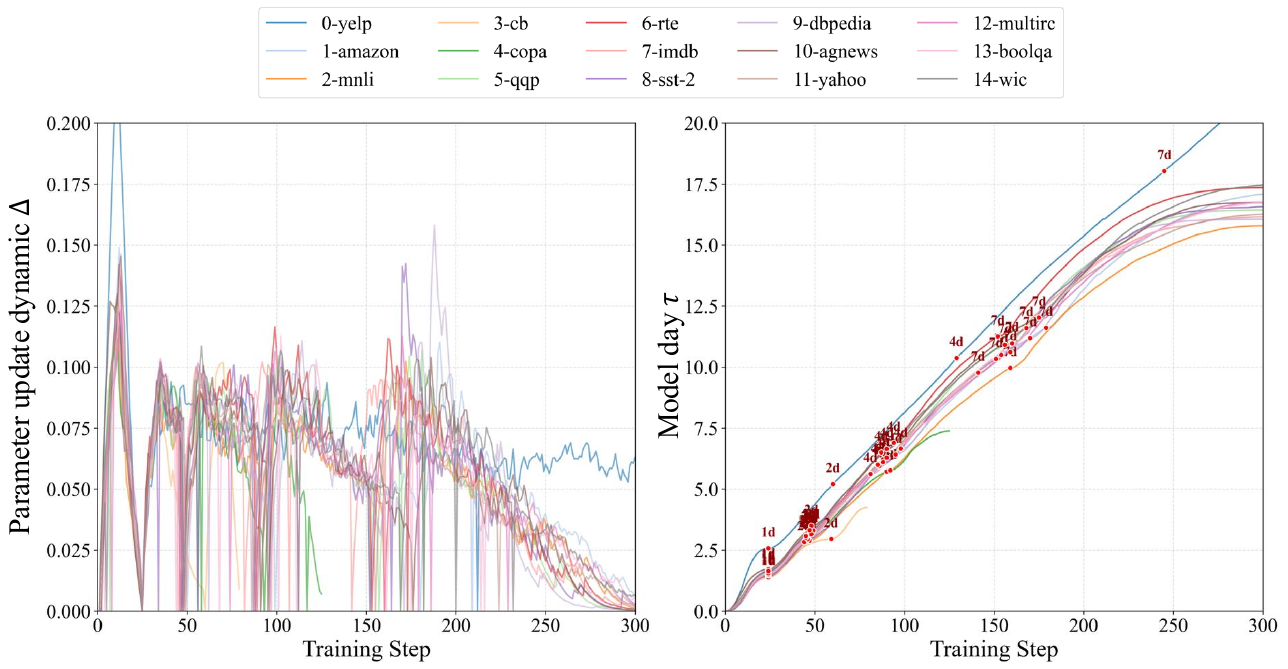}
        \caption{Long Sequence Benchmark, task order 6}
    \end{subfigure}

    \vspace{0.4em}

    \begin{subfigure}[t]{0.48\linewidth}
        \centering
        \includegraphics[width=\linewidth]{imgs/Vis_main.pdf}
        \caption{SuperNI Benchmark, task order 7}
    \end{subfigure}
    \hfill
    \begin{subfigure}[t]{0.48\linewidth}
        \centering
        \includegraphics[width=\linewidth]{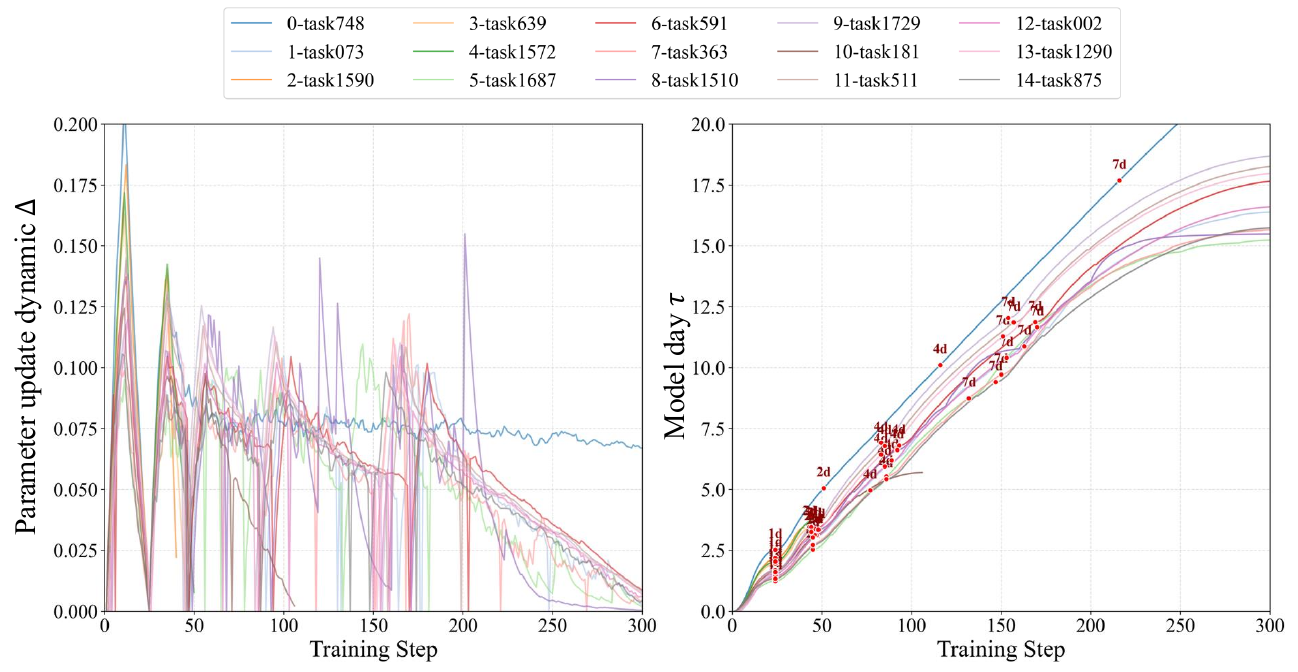}
        \caption{SuperNI Benchmark, task order 8}
    \end{subfigure}

    \caption{
\textbf{Replay dynamics across datasets and task orders.}
Each subfigure shows $\Delta_t$, accumulated model time $\tau_t$, and replay trigger points for one task order, illustrating adaptive replay scheduling based on model-centric time.
}

    \label{fig:8sub}
\end{figure*}

This appendix provides additional visualizations of replay dynamics for all task orders in the three benchmark (Fig.~\ref{fig:8sub}).
For each task order, we plot the step-wise parameter update magnitude $\Delta_t$, the accumulated model-centric time $\tau_t$, and the corresponding replay trigger points induced by {\ouralg}.

These visualizations further illustrate that parameter update dynamics vary substantially across tasks and training stages, leading to non-uniform growth of $\tau_t$.
As a result, replay events are triggered at different training steps for different tasks, even when following the same Ebbinghaus-inspired schedule.
This behavior highlights the adaptive nature of the proposed model-centric replay mechanism and contrasts with step-based replay strategies, which trigger replay at fixed iteration indices regardless of model state.

\section{Baselines}
\label{appendix:baseline}
We compare {\ouralg} with a comprehensive set of CL baselines, with a particular emphasis on replay–based methods that vary in their replay scheduling strategies. 
For fairness, all methods are implemented using the LoRA framework. 
(1) \textit{\textbf{MixReplay}}: Mixes samples from the current task and the memory buffer, and trains on the combined data at each step.
(2) \textit{\textbf{Fixed-interval Replay}}: Triggers replay at fixed, step-based intervals during training.
(3) \textit{\textbf{EWC}}~\cite{kirkpatrick2017overcoming}: Applies a regularization term to prevent interference with previously learned tasks.
(4) \textit{\textbf{O-LoRA}}~\cite{wang2023orthogonal}: Learns task-specific LoRA adapters in orthogonal subspaces.
(5) \textit{\textbf{MoELoRA}}~\cite{luo2024moelora}: Adopts the mixture-of-experts architecture for continual learning.
(6) \textit{\textbf{SAPT}}~\cite{zhao2024sapt}: Uses a shared-attention mechanism to align LoRA block selection across tasks.
(7) \textit{\textbf{MIGU}}~\cite{du2024unlocking}: Updates task-relevant parameters based on gradient magnitude to mitigate forgetting.
(8) \textit{\textbf{SSR}}~\cite{huang2024mitigating}: A self-synthesized rehearsal method that generates synthetic instances using the LLM for rehearsal. 
(9) \textit{\textbf{Recurrent-KIF}}~\cite{feng2025recurrent}: Performs model merging with dynamic parameter importance estimation.
(10) \textit{\textbf{AIMMerging}}~\cite{feng2025aimmerging}: Applies adaptive iterative model merging using training trajectories.
(11) \textit{\textbf{VBM}}~\cite{kang2025your}: A training step-based replay method inspired by the Ebbinghaus forgetting curve.
To ensure a fair comparison, all methods are trained for the same number of epochs.
Finally, \textit{\textbf{MTL}} (multi-task learning), which jointly trains on all tasks, serves as an upper-bound reference. 

\section{Dataset Statistics}
\label{sec:dataset}
We adopt the experimental setup from \citet{feng2024tasl2}, using three CL benchmark datasets:
(i) \textbf{Standard CL Benchmark}, which consists of five text classification tasks from \citet{zhang2015character}: AG News, Amazon Reviews, Yelp Reviews, DBpedia, and Yahoo Answers.
(ii) \textbf{Long Sequence Benchmark}, a more challenging evaluation scenario comprising 15 tasks \cite{razdaibiedina2023progressive}: five from the Standard CL Benchmark, four from the GLUE benchmark, five from SuperGLUE, and the IMDB Movie Reviews dataset.
(iii) \textbf{SuperNI Benchmark}~\cite{wang2022super}, a comprehensive benchmark designed to evaluate a wide range of NLP tasks, includes tasks in dialogue generation \cite{dong2024zero}, information extraction \cite{logicrag}, question answering \cite{zhao2024large}, summarization \cite{feng2024continual}, and sentiment analysis \cite{chen2024entity}.

Table \ref{superni} \& \ref{long-sequence} show details of the datasets we used for our experiments, along with their evaluation metrics. Overall, in SuperNI, we choose 3 tasks from dialogue generation (Dialog) \cite{chumodel, feng2023towards}, information extraction (IE) \cite{xu2025parenting},  question answering (QA), summarization (Sum) \cite{shi2024understanding} and sentiment analysis (SA) \cite{zhou2025dropping}, respectively.
For the Long Sequence benchmark, this includes five tasks from the standard CL benchmark (AG News, Amazon reviews, Yelp reviews, DBpedia and Yahoo Answers), four from GLUE benchmark (MNLI, QQP, RTE, SST2), five from SuperGLUE benchmark (WiC, CB, COPA, MultiRC, BoolQ), and the IMDB movie reviews dataset \cite{chustackelberg, hu-etal-2025-longrecipe, kang2025hssbenchbenchmarkinghumanitiessocial}.

We report 8 different task orders used for our experiments in Table \ref{order}.

\begin{table*}
\centering
\scalebox{0.9}{
\begin{tabular}{lllll}
\toprule
\textbf{Dataset name} & \textbf{Task}  & \textbf{Metric} \\
\midrule
1. task639\_multi\_woz\_user\_utterance\_generation  & dialogue generation   & Rouge-L        \\
2. task1590\_diplomacy\_text\_generation & dialogue generation   & Rouge-L       \\
3. task1729\_personachat\_generate\_next & dialogue generation   & Rouge-L      \\
4. task181\_outcome\_extraction & information extraction & Rouge-L        \\
5. task748\_glucose\_reverse\_cause\_event\_detection & information extraction & Rouge-L       \\
6. task1510\_evalution\_relation\_extraction   & information extraction & Rouge-L  \\
7. task002\_quoref\_answer\_generation & question answering & Rouge-L \\
8. task073\_commonsenseqa\_answer\_generation & question answering & Rouge-L    \\
9. task591\_sciq\_answer\_generation  & question answering & Rouge-L        \\
10. task511\_reddit\_tifu\_long\_text\_summarization     & summarization        & Rouge-L        \\
11. task1290\_xsum\_summarization  & summarization       & Rouge-L        \\
12. task1572\_samsum\_summary  &summarization  & Rouge-L \\
13. task363\_sst2\_polarity\_classification  & sentiment analysis   & accuracy        \\
14. task875\_emotion\_classification & sentiment analysis   & accuracy  \\
15. task1687\_sentiment140\_classification & sentiment analysis   & accuracy  \\
\bottomrule
\end{tabular}}
\caption{The details of 15 datasets in the SuperNI Benchmark \cite{wang2022super}.
}
\label{superni}
\end{table*}

\begin{table*}[htbp]
\centering
\scalebox{0.9}{
\begin{tabular}{lllll}
\toprule
\textbf{Dataset name} & \textbf{Category} & \textbf{Task}             & \textbf{Domain}     & \textbf{Metric} \\ \midrule
1. Yelp               & CL Benchmark      & sentiment analysis        & Yelp reviews        & accuracy        \\
2. Amazon             & CL Benchmark      & sentiment analysis        & Amazon reviews      & accuracy        \\
3. DBpedia            & CL Benchmark      & topic classification      & Wikipedia           & accuracy        \\
4. Yahoo              & CL Benchmark      & topic classification      & Yahoo Q\&A          & accuracy        \\
5. AG News            & CL Benchmark      & topic classification      & news                & accuracy        \\
6. MNLI               & GLUE              & natural language
inference                       & various             & accuracy        \\
7. QQP                & GLUE              & paragraph detection       & Quora               & accuracy        \\
8. RTE                & GLUE              & natural language inference                       & news, Wikipedia     & accuracy        \\
9. SST-2              & GLUE              & sentiment analysis        & movie reviews       & accuracy        \\
10. WiC               & SuperGLUE         & word sense disambiguation & lexical databases   & accuracy        \\
11. CB                & SuperGLUE         & natural language
inference                       & various             & accuracy        \\
12. COPA              & SuperGLUE         & question and answering                        & blogs, encyclopedia & accuracy        \\
13. BoolQA            & SuperGLUE         & boolean question and answering                & Wikipedia           & accuracy        \\
14. MultiRC           & SuperGLUE         & question and answering                        & various             & accuracy        \\
15. IMDB              & SuperGLUE         & sentiment analysis        & movie reviews       & accuracy        \\ \bottomrule
\end{tabular}}
\caption{The details of 15 classification datasets in the Long Sequence Benchmark \cite{razdaibiedina2022progressive}. First five tasks
correspond to the standard CL benchmark \cite{zhang2015character}.
}
\label{long-sequence}
\end{table*}

\begin{table*}[h]
\centering
\scalebox{0.9}{
\begin{tabular}{lll}
\hline
\textbf{Order} & \textbf{Benchmark} & \textbf{Task Sequence}                                                                                                                                \\ \hline
1              &  \multirow{3}*{\tabincell{c}{Standard CL}}      & dbpedia → amazon → yahoo → ag                                                                                                                         \\
2              &       & dbpedia → amazon → ag → yahoo                                                                                                                         \\
3              &       & yahoo → amazon → ag → dbpedia                                                                                                                         \\ \hline
4              & \multirow{3}*{\tabincell{c}{\\ Long Sequence}}              & \begin{tabular}[c]{@{}l@{}}mnli → cb → wic → copa → qqp → boolqa → rte → imdb →\\ yelp → amazon → sst-2 → dbpedia → ag → multirc → yahoo\end{tabular} \\
5              &             & \begin{tabular}[c]{@{}l@{}}multirc → boolqa → wic → mnli → cb → copa → qqp → rte\\ → imdb → sst-2 → dbpedia → ag → yelp → amazon → yahoo\end{tabular} \\
6              &              & \begin{tabular}[c]{@{}l@{}}yelp → amazon → mnli → cb → copa → qqp → rte → imdb →\\ sst-2 → dbpedia → ag → yahoo → multirc → boolqa → wic\end{tabular} \\ \hline
7              & \multirow{3}*{\tabincell{c}{\\ SuperNI}}  & \begin{tabular}[c]{@{}l@{}}task1572 → task363 → task1290 → task181 → task002 →\\ task1510 → task639 → task1729 → task073 → task1590 →\\ task748 → task511 → task591 → task1687 → task875\end{tabular} \\
8              &      & \begin{tabular}[c]{@{}l@{}}task748 → task073 → task1590 → task639 → task1572 →\\ task1687 → task591 → task363 → task1510 → task1729 →\\ task181 → task511 → task002 → task1290 → task875\end{tabular} \\ \hline
\end{tabular}}
\caption{Eight different orders of task sequences used for continual learning experiments. Orders 1-3 correspond to the standard CL becnhmark adopted by prior works. Orders 4-6 are long-sequence orders spanning 15 tasks, and orders 7-8 are superni spanning 15 tasks following \cite{razdaibiedina2023progressive}.}
\label{order}
\end{table*}

\section{Implementation Details}
\label{sec:details}

All experiments are implemented using PyTorch and the Transformers library, and conducted on 8 NVIDIA H20 GPUs. We use a learning rate of $3 \times 10^{-4}$ and a batch size of 8. Each new task is trained for 10 epochs, and when replay is triggered, previously learned tasks are trained for two additional epochs.

For LoRA, we set the rank $r=8$, scaling factor $\alpha=32$, and dropout rate to 0.05, and apply LoRA to the \texttt{q\_proj} and \texttt{v\_proj} modules. During inference, we use a temperature of 0.02, with top-$p$ sampling disabled, top-$k$=1, beam size set to 1, and a maximum of 128 newly generated tokens.

It is worth noting that we use the same hyperparameters across different datasets and backbones, demonstrating the generalizability of our method without requiring extensive hyperparameter tuning for each specific setting.
And all baselines are evaluated using identical model architectures, training configurations, and memory buffer sizes.

The detailed implementation of {\ouralg} is provided in Algorithm~\ref{alg:my_algorithm}.

\begin{algorithm*}[t]
\caption{\textbf{{\ouralg}}: \textbf{FOR}g\textbf{E}tting cur\textbf{V}e-inspired m\textbf{E}mory \textbf{R}eplay}
\label{alg:my_algorithm}
\begin{algorithmic}[1]
\REQUIRE 
Current task dataset $\mathcal{D}_k$; memory buffer $\mathcal{M}_{<k}$; model weights after training on task $k-1$, $\Theta^{k-1}$;
human replay intervals $\mathcal{D}_{\text{human}}$;
warm-up length $S$;
EMA coefficient $\lambda$;
scaling factor $\gamma$;
clipping bounds $g_{\min}, g_{\max}$;
base coefficient $\beta_{\text{base}}$.
\ENSURE Final parameters $\Theta^k$.

\STATE $\Theta^\star \gets \Theta^{k-1}$ \textit{\# anchor from previous task}
\STATE Initialize $\tau \gets 0$, $\mu \gets 0$, $j \gets 1$

\STATE \textit{\# Warm-up calibration}
\STATE Compute $\{\Delta_t\}_{t=1}^S$ on $\mathcal{D}_k$
\STATE $\tau_{\text{day}} \gets \sum_{t=1}^S \Delta_t$
\STATE $\mu_0 \gets \frac{1}{S}\sum_{t=1}^S \Delta_t$, $\mu \gets \mu_0$
\STATE $\mathcal{D}_{\text{model}} \gets \{ d \cdot \tau_{\text{day}} \mid d \in \mathcal{D}_{\text{human}}\}$

\STATE \textit{\# Training with scheduled replay}
\WHILE{training on $\mathcal{D}_k$}
    \STATE Update $\Theta$ on current task data
    \STATE Measure update $\Delta_t$; update $\tau \gets \tau + \Delta_t$
    \STATE Update $\mu \gets (1-\lambda)\mu + \lambda \Delta_t$

    \IF{$j \le |\mathcal{D}_{\text{model}}|$ \textbf{and} $\tau \ge \mathcal{D}_{\text{model}}[j]$ \textbf{and} $k>1$} 
        \STATE \textit{\# Replay begin}
        \STATE $r \gets \mu / (\mu_0 + \epsilon)$
        \STATE $s \gets \mathrm{clip}(1 + \gamma(r-1), g_{\min}, g_{\max})$
        \STATE $\beta \gets \beta_{\text{base}} \cdot s$
        \STATE Optimize replay loss on $\mathcal{M}$ for two replay epochs
        \STATE $j \gets j + 1$
    \ENDIF
\ENDWHILE

\STATE \textit{\# End-of-task consolidation}
\STATE Optimize replay loss on $\mathcal{M}$ for one short epoch via Eq. (\ref{eq:loss})

\STATE Update memory buffer $\mathcal{M}_{<k+1}$

\STATE \textbf{return} $\Theta^k$
\end{algorithmic}
\end{algorithm*}


\end{document}